\documentclass[10pt,twocolumn,letterpaper,table]{article}

\usepackage{iccv}              % To produce the CAMERA-READY version
\usepackage{bbding}
\usepackage{multicol}
\usepackage{multirow}
\usepackage{makecell}
\definecolor{iccvblue}{rgb}{0.21,0.49,0.74}
\usepackage[pagebackref,breaklinks,colorlinks,allcolors=iccvblue]{hyperref}
\usepackage{graphicx}

\def\paperID{-} % *** Enter the Paper ID here
\def\confName{ICCV}
\def\confYear{2025}

\title{SGA-INTERACT: 	
A 3D Skeleton-based Benchmark for Group Activity Understanding in Modern Basketball Tactic}

\author{Yuchen Yang\thanks{Work performed during his internship at Shanghai Artificial Intelligence Laboratory.} \textsuperscript{~1,2}
\quad Wang Wei\textsuperscript{2} 
\quad Yifei Liu\textsuperscript{3,2}
\quad Linfeng Dong\textsuperscript{4,2}
\quad Hao Wu\textsuperscript{5,2}
\quad Mingxin Zhang\textsuperscript{6} \\
\quad Zhihang Zhong\textsuperscript{2}
\quad Xiao Sun\textsuperscript{\Envelope~2}\\
\textsuperscript{1}Fudan University \quad \textsuperscript{2}Shanghai Artificial Intelligence Laboratory \quad \textsuperscript{3}Beihang University \\
\textsuperscript{4}Zhejiang University \quad 
\textsuperscript{5}University of Science and Technology of China \quad 
\textsuperscript{6}Shanghai University of Sport} %
\begin{document}

\maketitle

\begin{abstract}
Group Activity Understanding is predominantly studied as Group Activity Recognition (GAR) task.
However, existing GAR benchmarks suffer from coarse-grained activity vocabularies and the only data form in single-view, which hinder the evaluation of state-of-the-art algorithms.
To address these limitations, we introduce SGA-INTERACT, the first 3D skeleton-based benchmark for group activity understanding.
It features complex activities inspired by basketball tactics, emphasizing rich spatial interactions and long-term dependencies.
SGA-INTERACT introduces Temporal Group Activity Localization (TGAL) task, extending group activity understanding to untrimmed sequences, filling the gap left by GAR as a standalone task.
In addition to the benchmark, we propose One2Many, a novel framework that employs a pretrained 3D skeleton backbone for unified individual feature extraction.
This framework aligns with the feature extraction paradigm in RGB-based methods, enabling direct evaluation of RGB-based models on skeleton-based benchmarks.
We conduct extensive evaluations on SGA-INTERACT using two skeleton-based methods, three RGB-based methods, and a proposed baseline within the One2Many framework. 
The general low performance of baselines highlight the benchmark’s challenges, motivating advancements in group activity understanding.
The code is available at \url{https://github.com/Charrrrrlie/SGA-INTERACT}.
\end{abstract}
\section{Introduction}
Understanding group behavior for humans has emerged as an important problem in spatial-temporal modeling, requiring consideration of human-human and human-environment interactions over a time span.
It is primarily studied within Group Activity Recognition (GAR) in video analysis.
GAR aims at classifying the activity collectively performed by individuals in trimmed sequences \cite{wu2021comprehensive}. 
To advance GAR research, various benchmarks have been established across different domains, including daily life \cite{NUS-HGA,BEHAVE,UCLACourtyard,NURSINGHOME,CAD,CADE,CADN}, and sports \cite{BroadcastFieldHockey,NCAA,Volleyball,VolleyballTactic,CSport,NBA}.

\begin{figure}[t]
    \centering
    \includegraphics[width=\linewidth]{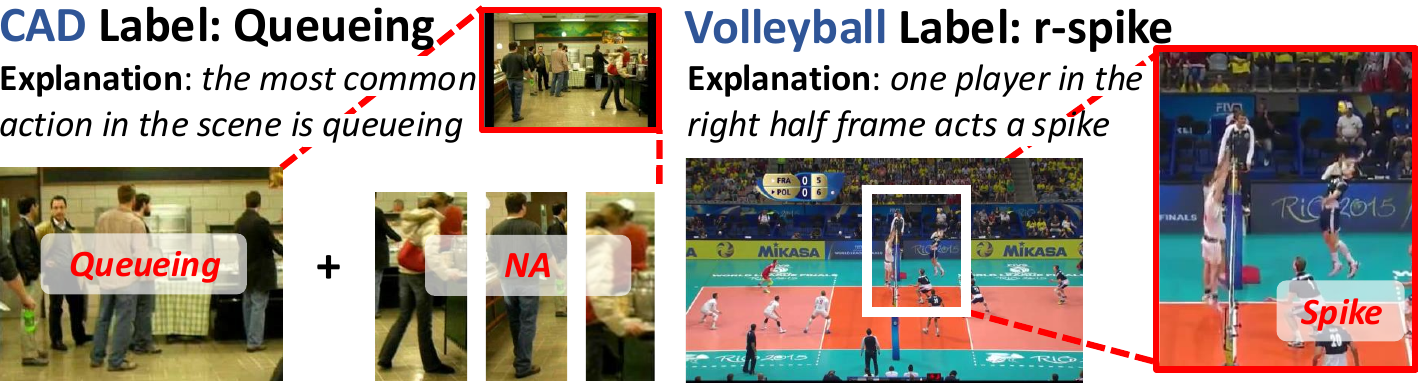}
    \caption{Overview of prevailing GAR datasets. Without much spatial and temporal dependencies, group activity can be recognized through a small number of individual actions in key frames.}
    \label{fig:existing-benchmarks}
\end{figure}

However, even explored for decades, the group activity understanding task still faces general challenges.
\textbf{1)} As illustrated in \cref{fig:existing-benchmarks}, current group activity labels exhibit limited semantic granularity, capturing only a few individual actions in a short moment. It makes the spatial-temporal reasoning capability of state-of-the-art methods under-evaluated.
\textbf{2)} Current group activity source data is limited to single-view RGB videos, making it sensitive to viewpoint changes. This sensitivity leads to appearance variations and occlusions of key information, reducing the robustness and generalization of models.
\textbf{3)} The GAR task remains limited in group activity understanding. It cannot cover the more practical scenario of multiple activities in one untrimmed sequence as input.
% \textbf{3)} With the advantages of skeleton representation, there is no benchmark specifically constructed for 3D skeletons. Compared to 2D representation, 3D skeleton disentangles occlusion/truncation from activity understanding and offers extra information from multi-view, advancing the research in crowd scene application.

\begin{table*}[t]
    \centering
 {\scriptsize
\captionof{table}{Comparison of prevailing GAR datasets on statistics. Clip length in frames. \# denotes the number of one index.}
    \begin{tabular}{l|c|c|c|c|c|c|c|c|c|c}
    \toprule
    {} & \multicolumn{7}{c}{Source Data} & \multicolumn{3}{|c}{Annotation} \\
    \cmidrule{2-11}
    {} & {Multi-view} & {Source} & {\#Video} & {\#Clip} & {FPS} & {Resolution} & {\#Length} & {Trajectory} & {Boundary} & {\#Activity} \\
    \midrule
    {CAD~\cite{CAD}} & {\XSolidBrush} & {Handheld Camera} & {$44$} & {$2,511$} & {-} & {$480p$} & {$10$} & {\Checkmark} & {\XSolidBrush} & {$5$}
    \\
    {Volleyball~\cite{Volleyball}} & {\XSolidBrush} & {Youtube} & {$55$} & {$4,830$} & {-} & {$720p$} & {$40$} & {\Checkmark} & {\XSolidBrush}
    & {$8$} \\
    {NBA~\cite{NBA}} & {\XSolidBrush} & {Youtube} & {$181$} & {$9,172$} & {$12$} & {$720p$} & {$72$} & {\XSolidBrush} & {\XSolidBrush} & {$9$}
    \\
    \midrule
    {SGA-INTERACT} & {\Checkmark} & {MoCAP} & {$95\times4$} & {$3,120\times4$} & {$50$} & {$1080p$} & {$30\sim400$} & {\Checkmark} & {\Checkmark} & {$18$}\\
    \bottomrule
  \end{tabular}
  \label{table:benchmark}
}
\end{table*}

From data modality perspective, skeleton sequences for group activity recognition have been gaining increasing attention.
Representing individuals using skeletons eliminates redundant background information and variations in illumination and textures presented in RGB videos, while also being less memory-consuming~\cite{sun2022human}.
% Researchers employ off-the-shelf pose estimators to extract 2D skeletons from the existing datasets, enabling the development of skeleton-based methods \cite{GRIN,POGARS,ZoomTransformer,COMPOSER,zhang2024bi,MPGCN} achieving comparable performance with RGB-based ones.
By employing off-the-shelf 2D pose estimators to extract skeletons from the existing dataset, researchers develop skeleton-based methods \cite{GRIN,POGARS,ZoomTransformer,COMPOSER,zhang2024bi,MPGCN}, 
which achieves comparable performance with RGB-based ones.
For benchmarking, skeleton sequences avoid some potential privacy and ethical issues, making it convenient for data collection.

Based on the limitations of existing benchmarks and the advantage of skeleton representation, we propose \textbf{SGA-INTERACT} (3D \textbf{S}keleton-based dataset for \textbf{G}roup \textbf{A}ct\textbf{I}vity u\textbf{N}ders\textbf{T}anding in
mod\textbf{ER}n basketball \textbf{TACT}ic). The overview is presented in \cref{fig:teaser}.
We construct an accurate multi-view Motion Capture (MoCAP) environment to collect data from elite $3\times3$ basketball games of the Olympic Qualifying Tournament. Incorporated with professional basketball video analysts, we design and annotate atomic group activities derived from modern basketball tactics, describing complex but systematic movement of players with definitive start and end signs of activity boundary.
Statistically, as summarized in \cref{table:benchmark}, SGA-INTERACT possesses high spatio-temporal quality and long-term correlation, forming the first 3D skeleton-based group activity dataset with challenging labels.

As a complement to GAR, we introduce a new task, Temporal Group Activity Localization (TGAL), designed to assess the group activity understanding ability in untrimmed videos. It aims at temporally localizing group activity within a long video and classifying each localized activity into predefined categories.
Basketball rules, where each round follows a consistent format, provide a practical basis for TGAL algorithms. For TGAL, skeleton sequences from game rounds serve as input, with group activity boundaries and categories as ground truth.

To bridge RGB- and skeleton-based research, we introduce One2Many, a unified framework for both GAR and TGAL tasks.
Both RGB and skeleton-based methods share a process for extracting individual features and modeling group representation from individuals. 
In One2Many framework, we adopt pretrained ST-GCN~\cite{yan2018spatial}, a well-established human action recognition network for individual feature extraction. 
It leverages individual action recognition priors (``One'') for group activity understanding (``Many'').
The framework aligns with the feature extraction paradigm in RGB-based methods while taking skeleton sequences as input.
Thus, existing methods can be evaluated in our skeleton-based benchmark, focusing on the critical capability of group representation modeling.

In evaluation, we clarify the experimental settings in baseline methods and conduct detailed experiments to seek future research insights into group activity understanding.
Two open-sourced skeleton-based methods, three RGB-based methods, and a proposed method within the One2Many framework are implemented.
The generally low performance on GAR and TGAL tasks highlights the challenges posed by SGA-INTERACT benchmark, indicating that spatial temporal modeling capabilities have not yet reached the bottleneck.
Ablation studies demonstrate the benefits of extra information and 3D data, which supports SGA-INTERACT as a comprehensive benchmark for research.
Finally, experiments on the pretraining in One2Many framework reinforce the effectiveness of leveraging large-scale action recognition priors, motivating the development of backbone for group activity understanding.

Our contributions are summarized in three-fold:
\begin{itemize}
    \item We propose SGA-INTERACT, the first 3D skeleton-based group activity benchmark with new TGAL task, featuring rich interaction and long-term dependencies.
    \item We introduce One2Many, a unified framework for group activity understanding. It enables feasible evaluation for RGB-based methods on the skeleton-based benchmark.
    \item We conduct detailed evaluation of existing methods, providing valuable research directions for the community.
\end{itemize}
\section{Related Work}
\subsection{Group Activity Recognition Datasets}
Despite numerous group activity recognition datasets~\cite{NUS-HGA,BEHAVE,UCLACourtyard,NURSINGHOME,CAD,CADE,CADN,BroadcastFieldHockey,NCAA,Volleyball,VolleyballTactic,CSport,NBA} have been constructed, their widespread use has been limited by issues related to accessibility and data organization.
Among them, CAD~\cite{CAD}, Volleyball~\cite{Volleyball}
and NBA~\cite{NBA} are the most commonly utilized. 
CAD assigns the largest number of individual actions within the scene as the group activity. Volleyball designs four basic activities and differentiates them by left and right court. 
As illustrated in \cref{fig:existing-benchmarks} and \cref{table:benchmark},
in the spatial dimension, they lack interaction, which diminishes the meaning of the term ``group''.
In the temporal dimension, CAD and Volleyball annotate activities in sampled frames by a fixed stride and assume a certain range of neighboring frames as the activity boundary.
They overlook the actual activity boundary, interfering with recognition by involving inconsistent patterns.
Furthermore, only short-term dependencies are involved.
NBA also suffers from these drawbacks. Moreover, NBA does not provide human tracking annotations, which harms group activity understanding that requires high spatial-temporal correlation.

Recently, social group activity recognition~\cite{ehsanpour2020joint,ehsanpour2022jrdb,tamura2022hunting,kim2024towards} has been studied in the situation of multiple groups within a single scene, easing the limitation of GAR. However, group activity understanding in untrimmed sequences remains unexplored.

\subsection{Group Activity Recognition Methods}
Group Activity Recognition (GAR) has been comprehensively studied in the past decades.
According to the modality of input, GAR methods are categorized into RGB-based~\cite{CRM,ARG,GroupFormer,DualAI,li2022learning,xie2023actor}, skeleton-based~\cite{GRIN,POGARS,ZoomTransformer,COMPOSER,zhang2024bi,MPGCN}, and RGB-skeleton fusion~\cite{AT,SACRF,GroupFormer,pei2023key} methods. In addition, several works~\cite{CRM,AT,GRIN,GroupFormer,SACRF,DualAI,xie2023actor} fuse optical flow to further enhance group representation. 
% Those methods focus on modeling individual relations in both spatial (human-to-human, human-to-object) and temporal dimensions.
From the perspective of deep learning components, Graph Neural Networks (GNNs) are employed in \cite{ARG, DIN, xie2024active, MPGCN} to leverage the effectiveness of graph-based topology modeling.
Several methods~\cite{AT,GroupFormer,COMPOSER,pei2023key} utilize attention~\cite{vaswani2017attention} for its strong relation modeling ability. Causality~\cite{yuan2021learning,xie2023actor,zhang2024bi} and Conditional Random Field~\cite{SACRF} are introduced to improve reasoning performance.
Although existing methods achieve competitive results in prevailing GAR datasets, limited by the datasets themselves, the ability in spatial-temporal modeling of existing methods remains poorly evaluated.

\section{SGA-INTERACT}
Sports scenes involve rapid movement and dynamic interactions among groups of people, where tactics reveal complex semantic patterns in systematic behaviors~\cite{wang2024tacticai,li2021multisports}. Drawing on these characteristics, SGA-INTERACT aims to establish a challenging benchmark for group activity understanding by collecting high-quality basketball tactic sequences and comprehensive annotations.

\subsection{Dataset Construction}
\label{sec:data-cons}
\noindent
\textbf{Raw Data Collection.} 
To obtain sport tactic data in high quality for group activity understanding, we focus on elite $3\times3$ basketball games from the 2024 Olympic Qualifying Tournament, which involves 12 professional teams and consists of 100 games across 5 locations over a month.
We construct a multi-view Motion Capture (MoCAP) environment for data collection.
Four consumer-grade cameras are positioned at each corner of the basketball court, all operating at $1920\times1080$ ($1080p$) resolution and $50$ FPS to capture motion details. Camera intrinsic and extrinsic parameters are calibrated by leveraging checkerboard~\cite{zhang2002flexible} and keypoints of the basketball court, respectively.

\begin{figure}[t]
    \centering
    \includegraphics[width=\linewidth]{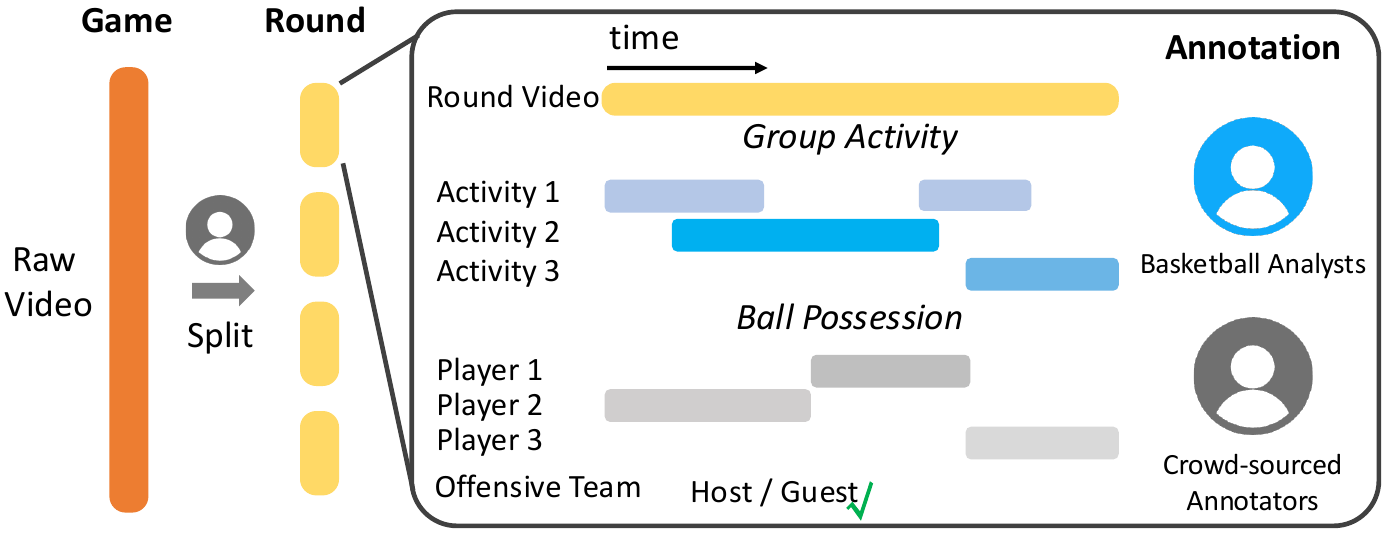}
    \caption{Overview of SGA-INTERACT annotation pipeline.}
    \label{fig:annot-pipeline}
\end{figure}

\noindent
\textbf{Group Activity Vocabulary Construction.}
In the modern $3\times3$ basketball tactic system, each tactic involves a prescribed combination of movements, with athletes making adjustments based on the immediate situation.
% and various extensions by athletes according to the immediate situation. 
However, there is a gap between the formal definition of tactics and the desired vocabulary for group activity understanding. Ignoring the dynamics of tactics can result in chaotic patterns of recognition while distinguishing between various extensions leads to an overly complex categorization.
To bridge this gap, incorporated with professional basketball video analysts, we start with atomic movements and define 21 tactical movements as group activity vocabulary from the basketball tactic system. Each defined movement is distinct, with clear start and end points, making them ideal as group activities. In practice, we focus on 18 of these tactical movements, based on their frequency of occurrence.

\begin{figure*}[t]
    \centering
    \begin{subfigure}[t]{0.48\textwidth}
    {
        \centering
        \includegraphics[width=\textwidth]{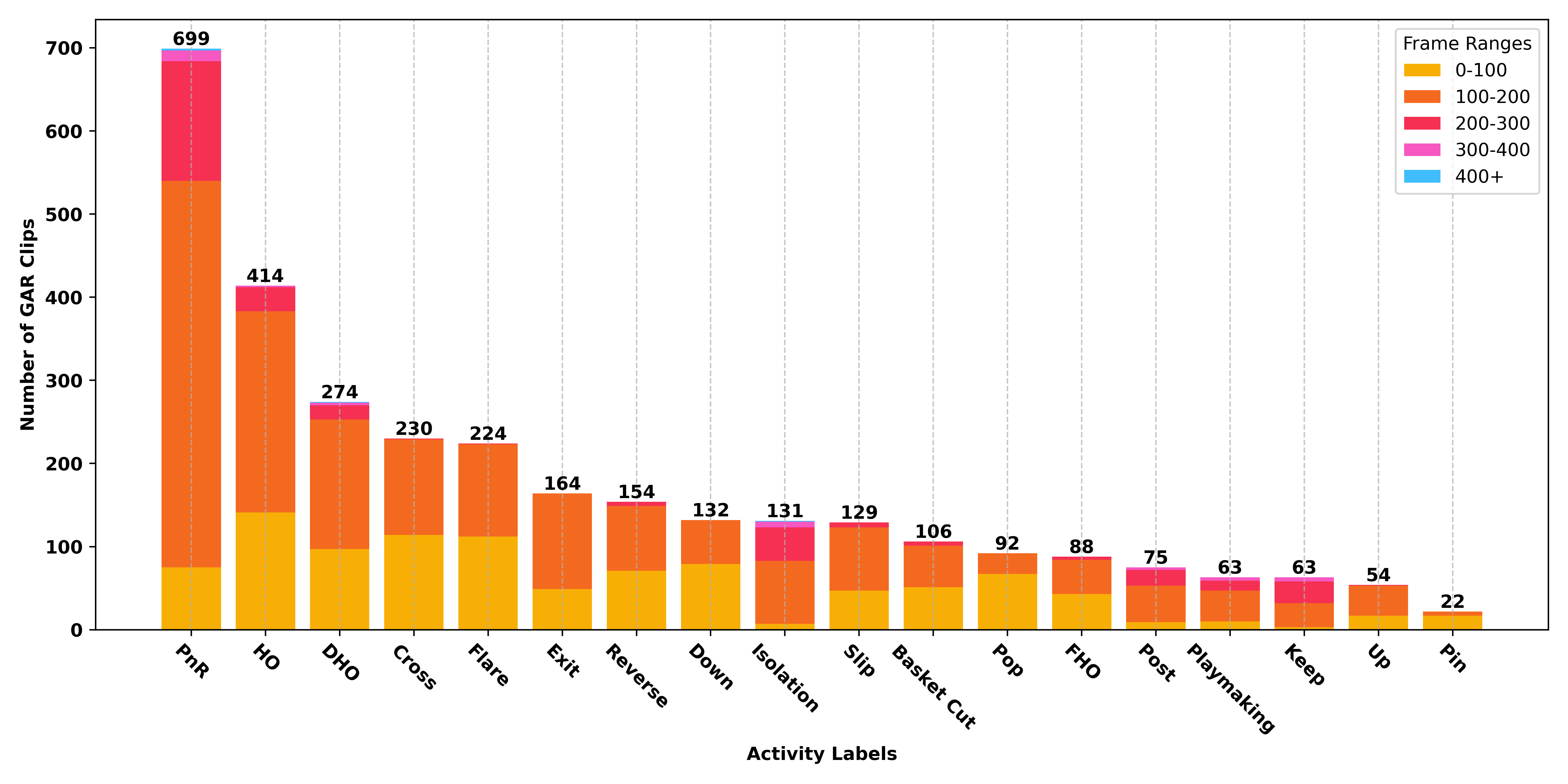}
        \caption{Number of GAR clips of various group activities. Clip length is distinguished by color.}
        \label{fig:stats-GAR}
    }
    \end{subfigure} \hfill
    \begin{subfigure}[t]{0.48\textwidth}
    {
        \centering
        \includegraphics[width=\textwidth]{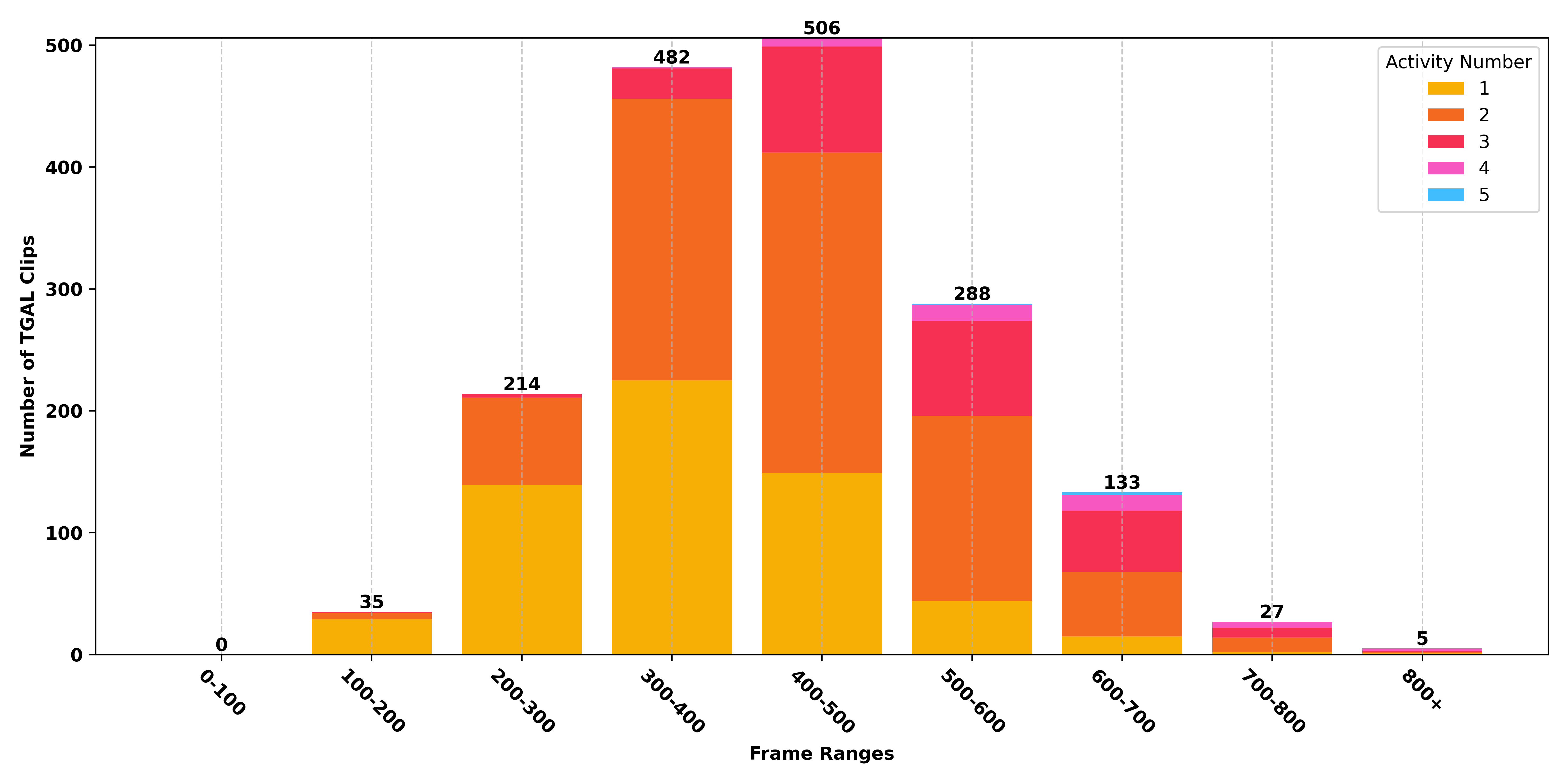}
        \caption{Number of TGAL clips of various clip lengths. The number of group activities within a clip is distinguished by color.}
        \label{fig:stats-TGAL}
    }
    \end{subfigure} \hfill
    \caption{Statistics of SGA-INTERACT dataset for GAR and TGAL tasks.}
    \label{fig:stats}
\end{figure*}

\noindent
\textbf{Data Annotation.}
Except for round segments, each raw video of a game also includes irrelevant segments of timeout, substitution, etc.
To efficiently annotate raw video data, we develop a multi-stage annotation pipeline. 
As illustrated in \cref{fig:annot-pipeline}, a group of crowd-sourced annotators with limited basketball expertise first identify and record the start and end frame of each round from raw videos. As the important information for action recognition validated in \cite{Volleyball}, ball information and offensive team information during each round are also annotated in the SGA-INTERACT dataset by those annotators.
Instead of ball location, which implicitly provides human-object relation, we annotate the ball carrier to provide more direct ball possession information.
The annotation is based on the VIG online annotation tool~\cite{VGGAnnot} for temporal accuracy.
For group activity labels, we employ a team of professional basketball video analysts with over three years of experience working with elite basketball clubs. The analysts use Sportscode Elite~\cite{Sportscode}, a software they are highly proficient in, to annotate group activity categories and temporal boundaries (start and end frames).

\noindent
\textbf{3D Skeleton Data Collection.}
We acquire 3D pose sequences of the 6 players on the court by multi-person multi-view 2D pose estimation and 3D pose reconstruction in a top-down strategy.
For each camera view, we employ RTMDet~\cite{lyu2022rtmdet} and RTMPose~\cite{jiang2023rtmpose} to detect players and estimate keypoints in the COCO format~\cite{lin2014microsoft}. Unlike previous methods~\cite{GRIN,POGARS,COMPOSER}, utilizing OpenPose~\cite{cao2017realtime}, Hourglass~\cite{newell2016stacked} or HRNet~\cite{sun2019deep} for pose estimation, the advanced RTMPose provides more accurate results.
A database of 200 profile images per player is created for the subsequent tracking and re-identification (Re-ID). 
Note that the database records the team information of players, once player trajectories are matched, the extra annotation of team information is provided.
In practice, we employ OCSORT tracker~\cite{cao2023observation} and SOLIDER Re-ID algorithm~\cite{chen2023beyond} incorporating 3D matching penalty during the matching process.
After obtaining the pose trajectories of each player, we apply Triangulation~\cite{hartley2003multiple} to reconstruct the final 3D pose sequences.
Based on the basketball court size, we filter out the results that fall outside the court boundaries.
Finally, we review all the skeleton sequences and manually correct the wrong matching cases.

\noindent
Refer to \cref{sec:data-cons-detail} for dataset construction details.

\noindent
\textbf{Quality Control.}
For data annotation, both crowd-sourced annotators and professional basketball analysts follow a structured process, consisting of pre-annotation, formal annotation, and cross-check stages. In the pre-annotation stage, we select $1\%$ of the data to familiarize annotators with operations and ensure consistency before formal annotation begins. 
Each annotated item is then reviewed by a second annotator, who has not worked on that specific item, as part of the cross-checking process.
Additionally, 3D skeleton sequences are also double-checked.

\subsection{Task and Metrics}
As outlined in \cref{sec:data-cons}, SGA-INTERACT contains 3D skeleton sequences of game rounds and corresponding group activity labels with temporal boundaries.
It facilitates two research tasks:

\noindent
\textbf{TGAL.}
We introduce a new task, Temporal Group Activity Localization (TGAL), inspecting the ability of group activity understanding in untrimmed sequences.
Given a 3D skeleton sequence of a game round, denoted as untrimmed data $\mathbf{X}^{TGAL}$ with $T$ frames, the goal of TGAL is to localize a set of group activity instances $\mathbf{G}=\{\{t_m^s, t_m^e, c_m\}\}^M$ of number $M$, where $t_m^s$, $t_m^e$, $c_m$ denote start frame, end frame and category of the $m^{th}$ group activity. 

The evaluation metric is the average mean Average Precision (mAP) across all categories, computed under temporal Intersection over Union (tIoU).

The TGAL task shares similarities with the classical Temporal Action Localization (TAL) task in video understanding~\cite{idrees2017thumos, caba2015activitynet}, while the TGAL focuses on group behaviors. 
SoccerNet-v3~\cite{cioppa2022scaling} proposes the Action Spotting task, aiming at localizing a single frame that presents the most salient moment of activity. The label of Action Spotting is sparse and susceptible to annotator bias. TGAL avoids this issue by complete temporal boundary annotations.

\begin{figure*}[t]
    \centering
    \includegraphics[width=0.93\linewidth]{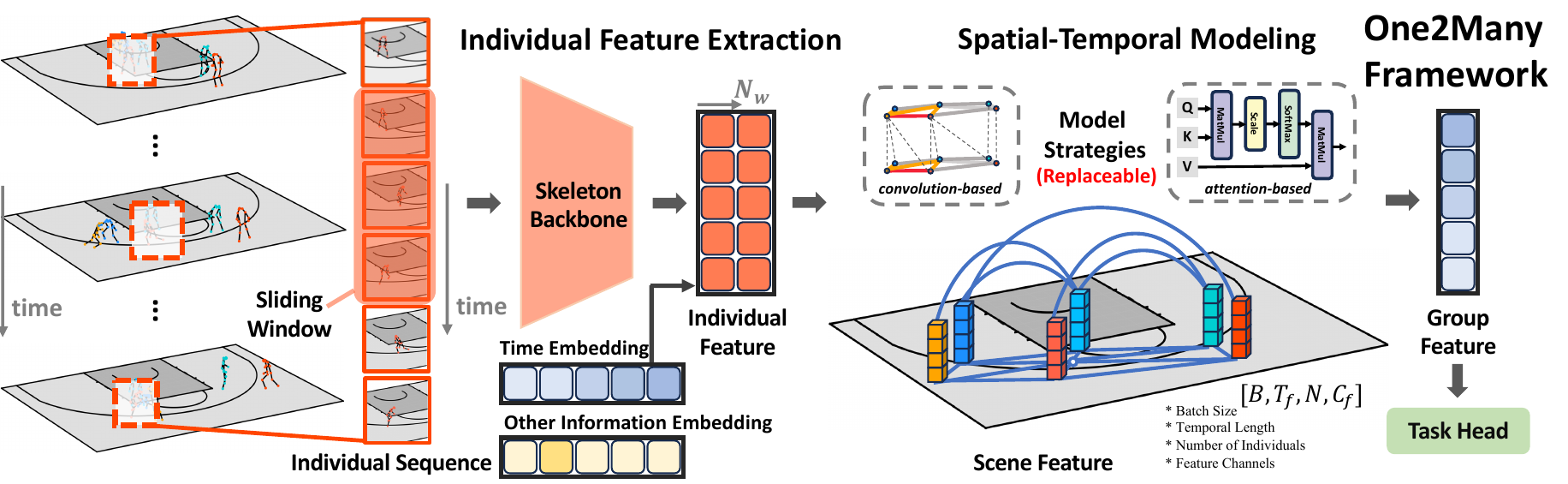}
    \caption{Overview of One2Many framework.}
    \label{fig:one2many}
\end{figure*}

\noindent
\textbf{GAR.} We restate the definition of Group Activity Recognition (GAR) and clarify its implementation in SGA-INTERACT. Given the temporal boundaries of multiple activities in game rounds, we trim the 3D skeleton sequences $\mathbf{X}^{TGAL}$ into segments $\mathbf{X}^{GAR}$, where each corresponds to a single group activity instance $\{c_m\}$ with $t_m^e - t_m^s + 1$ frames.
GAR aims at classifying each trimmed sequence $\mathbf{X}^{GAR}$ into one of the pre-defined categories.

For evaluation, previous methods mixed average accuracy $mAcc.$ and overall accuracy $oAcc.$ metrics, causing inconsistency of performance comparison.
$mAcc.$ calculates the accuracy for each category and averages it across all categories, while $oAcc.$ measures the overall accuracy across all items, regardless of category. 
Therefore, $oAcc.$ is sensitive to sample imbalance. 
We report both $mAcc.$ and $oAcc.$ for comprehensive evaluation. Additionally, Top$3$-$mAcc.$ is also applied for lenient measurement.

\subsection{Dataset Statistics}
SGA-INTERACT collects $\sim6$ million frames of multi-view raw data from $95$ elite 3$\times$ 3 basketball games. 
To support group activity understanding tasks, according to temporal annotations, SGA-INTERACT extracts $1,690$ game rounds of $\sim0.8$ million frames. 

For TGAL, a game round corresponds to one input sequence, resulting in $1,690$ TGAL clips.
For GAR, tactical sequences are trimmed from game rounds, producing $3,120$ GAR clips.
As shown in \cref{fig:stats-GAR}, the number of GAR clips varies from 22 to 699 (3 categories are removed by extremely low frequency as discussed in vocabulary construction). 
It presents long-tailed distribution in real application scenarios, raising challenges for both group activity recognition and localization.
As demonstrated in \cref{fig:stats-TGAL}, TGAL clips vary in sequence length and the number of contained group activities, 
providing graded difficulties in our dataset.
In \cref{sec:supp-data-dist}, we visualize 3D skeleton sequences to represent data distribution, showing the diversity and recognizable patterns of the designed group activities.

SGA-INTERACT splits training and testing sets by group activity categories with a $7:3$ ratio. We fix the splits for a fair comparison of follow-up methods.

\subsection{Dataset Characteristics}
SGA-INTERACT possesses several distinguishing characteristics compared to previous methods in \cref{table:benchmark}:

\noindent
\textbf{New data type with high quality.} SGA-INTERACT introduces the first 3D skeleton-based dataset for group activity understanding at a large scale. 
    3D sequences are collected by a high-precision MoCAP environment, equipped with advanced perception algorithms, featuring high spatial-temporal resolution of $50$ FPS and $1080p$. 

\noindent
\textbf{High semantic level group activity vocabulary in rich spatial interaction and long-term temporal dependency.} The dataset encompasses 18 meticulously designed activity categories, representing intricate tactical movements of multiple individuals. Meanwhile, $93.6\%$ of GAR clips extend beyond $72$ frames, significantly surpassing previous methods in long-term reasoning requirements.

\noindent
\textbf{Comprehensive and precise annotations.} Except for group activity category labels, SGA-INTERACT includes temporal boundaries of group activity, facilitating the novel TGAL task. Benefiting to annotators' expertise and strict quality control process, the dataset maintains a high standard of annotation accuracy and reliability.
\section{One2Many Framework}
The primary motivation behind the One2Many framework is to unify RGB- and skeleton-based methods by
leveraging priors from large scale data for group activity understanding.
One2Many employs a pretrained skeleton-based backbone, originally trained for individual action recognition, to extract individual features.
Then the individual features are aggregated and refined in spatial-temporal modeling as group features for GAR and TGAL tasks.

\subsection{Architecture}
For clarity, as shown in \cref{fig:one2many}, we summarize the group activity understanding task in three stages: individual feature extraction, spatial-temporal modeling for group feature extraction and task-specific output.
We introduce the One2Many design of each stage in turn.

\noindent
\textbf{Individual Feature Extraction.}
Group skeleton sequences $\mathbf{X} \in \mathbb{R}^{T\times N\times C}$ are given, where $T$, $N$, $C$ denotes the time length, number of individuals in the group and skeleton dimension, respectively.

Ball possession for each player is represented as a binary vector of length $T$, where moments of carrying the ball are marked as 1, and all other moments are marked as 0.
Team information is represented as a binary vector of length $N$, where players in the offensive team are set as 1, and players in the defensive team are set as 0.

One2Many employs ST-GCN~\cite{yan2018spatial} pretrained on NTU RGB+D~\cite{shahroudy2016ntu} dataset as the skeleton backbone $\mathbf{\mathcal{B}}$ to extract individual features.
Considering the sequence length is larger than the backbone input,
we utilize a sliding window with $W_l$ length and $W_s$ stride to extract local features in the skeleton sequences.
The feature $\mathbf{\mathcal{F}}_{i}^{w}$ of the $i^{th}$ individual in the $w^{th}$ sliding widow is represented as:
\begin{equation}
    \mathbf{\mathcal{F}}_{i}^{w} = \mathbf{\mathcal{B}}(\mathbf{X}_{[w*W_s:w*W_s+W_l,i,:]}).
\end{equation}
Then we adopt embeddings to incorporate temporal information into models. For time and ball possession, we use the Learned Absolute Positional Encoding~\cite{vaswani2017attention} to map each moment to a vector. 
For team information, we repeat it along time dimension before applying embedding.

After separately operating each individual data in all $ N_{w}$ sliding windows, the scene feature is obtained by aggregation:
\begin{equation}
    \mathbf{\mathcal{F}} = Agg(\{\mathcal{F}_{i}^{w}|i=1,...,N;w=0,...,N_{w}-1\}),
\end{equation}
where $\mathbf{\mathcal{F}} \in \mathbb{R}^{T_f \times N \times C_f}$. $T_f$ denotes the down-sampled $T$ from skeleton backbone. $C_f$ is the number of feature channel.
For GAR, we utilize concatenation for $Agg$, producing $N_w$ times of skeleton backbone features for each individual. For TGAL, average operation is adopted to saving memory from larger $N_w$.

\noindent
\textbf{\textit{Discussion:}}
Previous fusion-based GAR methods~\cite{AT,SACRF,GroupFormer} utilizes off-the-shelf 2D pose estimators~\cite{alphapose,sun2019deep} to extract individual skeleton-related features. 
However, since the extracted features source from single frame, these features lack temporal correlation and troubles RGB-based method evaluation on skeleton-based benchmarks.

The pretrained skeleton-based backbone $\mathbf{\mathcal{B}}$ in One2Many framework mirrors the process of image/video backbone~\cite{simonyan2014very,carreira2017quo,szegedy2016rethinking} with RoIAlign~\cite{he2017mask} in RGB-based methods. Therefore, One2Many enables a range of valuable RGB-based methods evaluated in the SGA-INTERACT benchmark, with subsequent modules unchanged.
Moreover, it eliminates the laborious hand-craft feature selection (i.e. velocity, acceleration) in existing skeleton-based methods. Instead, One2Many provides a unified representation, focusing on the spatial-temporal modeling stage.

\noindent
\textbf{Spatial-temporal Modeling.}
With the scene feature $\mathbf{\mathcal{F}}$ extracted in a structured form, various spatial-temporal modeling strategies can be integrated. In practice, we implement methods with distinct modeling stages as baselines.

Inspired by~\cite{DualAI,zhang2024bi}, we propose an attention-based modeling strategy, named STAtt, as a supplement. 
STAtt consists of $N_{ST}$ basic STBlocks, each designed to capture both spatial and temporal correlations of the scene feature. 
Specifically, the scene feature $\mathbf{\mathcal{F}}$ is first independently enhanced by self-attention~\cite{vaswani2017attention} along spatial and temporal dimension, respectively.
The enhanced features are then correlated in spatial-temporal and temporal-spatial aspects by cross-attention~\cite{vaswani2017attention}. Details are provided in~\cref{sec:supp-statt-strucutre}.

% Specifically, flattening $\mathbf{\mathcal{F}}$ spatial and temporal dimension into batch dimension respectively to obtain $\mathbf{\mathcal{F}}_{temp}$ and $\mathbf{\mathcal{F}}_{spatio}$. 

\noindent
\textbf{Task Head.}
Up to this point, the One2Many framework produces task-agnotisc group features.
To adapt to the GAR and TGAL tasks, we introduce task-specific heads for each output.
For GAR, we follow established methods to reduce the spatial and temporal dimensions, then use a multi-layer perceptron (MLP) for classification.
For TGAL, we leverage the head from ActionFormer~\cite{zhang2022actionformer}, which retains the temporal dimension and generates separate classification and regression heatmaps. The TGAL results are then decoded using the approach from CenterNet~\cite{zhou2019objects}.
More details are provided in \cref{sec:one2many-head}.

\subsection{Loss}
For classification in both GAR and TGAL, we use Focal Loss~\cite{lin2017focal} to address class imbalance.
For regression in TGAL, we employ Distance-IoU Loss~\cite{zheng2020distance} for precise boundary regression.

\begin{table*}[t]
    \centering
 {\scriptsize
\captionof{table}{Quantitative results of GAR task on SGA-INTERACT. $^\dagger$ indicates the bug-fixed version (verified by the author). The settings without any augmentation are distinguished in gray. The best performances with/without augmentation are highlighted.}
    \begin{tabular}{l|cc|c|c|c|c|c}
    \toprule
    {Method} & {Spatial Aug.} & {Temporal Aug.} & {mAcc.-\%} & {Top3-mAcc.-\%} & {oAcc.-\%} & {\#Param.} & {FLOPs} \\
    \midrule
    {\multirow{4}{*}{COMPOSER$^\dagger$~\cite{COMPOSER}}} & {} & {} & {\cellcolor{gray!40}$47.72$} & {\cellcolor{gray!40}$76.43$} & {\cellcolor{gray!40}$58.64$} & {\multirow{4}{*}{$20.14$M}} & {\multirow{4}{*}{$0.8$G}} \\
    {} & {\Checkmark} & {} & {$54.06$} & {$80.71$} & {$62.82$} & {} & {} \\
    {} & {} & {\Checkmark} & {$43.38$} & {$69.72$} & {$54.32$} & {} & {} \\
    {} & {\Checkmark} & {\Checkmark} & {$52.23$} & {$77.36$} & {$61.31$} & {} & {} \\

    \midrule
    {\multirow{4}{*}{MPGCN~\cite{MPGCN}}} & {} & {} & {\cellcolor{gray!40}$56.42$} & {\cellcolor{gray!40}$80.69$} & {\cellcolor{gray!40}$67.78$} & {\multirow{4}{*}{$2.25$M}} & {\multirow{4}{*}{$37.0$G}} \\
    {} & {\Checkmark} & {} & {$66.68$} & {$88.83$} & {$72.74$} & {} & {} \\
    {} & {} & {\Checkmark} & {$60.45$} & {$85.43$} & {$68.36$} & {} & {} \\
    {} & {\Checkmark} & {\Checkmark} & {$62.83$} & {$86.79$} & {$69.94$} & {} & {} \\

    \midrule
    {\multirow{4}{*}{One2Many-ARG~\cite{ARG}}} & {} & {} & {\cellcolor{gray!40}$37.98$} & {\cellcolor{gray!40}$65.03$} & {\cellcolor{gray!40}$52.59$} & {\multirow{4}{*}{$43.68$M}} & {\multirow{4}{*}{$117.3$G}} \\
    {} & {\Checkmark} & {} & {$59.42$} & {$85.57$} & {$66.37$} & {} & {} \\
    {} & {} & {\Checkmark} & {$50.47$} & {$80.70$} & {$62.07$} & {} & {} \\
    {} & {\Checkmark} & {\Checkmark} & {$54.67$} & {$80.51$} & {$64.22$} & {} & {} \\

    \midrule
    {\multirow{4}{*}{One2Many-AT~\cite{AT}}} & {} & {} & {\cellcolor{gray!40}$53.50$} & {\cellcolor{gray!40}$81.12$} & {\cellcolor{gray!40}$65.63$} & {\multirow{4}{*}{$9.01$M}} & {\multirow{4}{*}{$104.2$G}} \\
    {} & {\Checkmark} & {} & {$60.73$} & {$85.75$} & {$68.97$} & {} & {} \\
    {} & {} & {\Checkmark} & {$60.23$} & {$85.79$} & {$67.13$} & {} & {} \\
    {} & {\Checkmark} & {\Checkmark} & {$65.53$} & {$87.36$} & {$71.12$} & {} & {} \\

    \midrule
    {\multirow{4}{*}{One2Many-DIN~\cite{DIN}}} & {} & {} & {\cellcolor{gray!40}$57.31$} & {\cellcolor{gray!40}$84.21$} & {\cellcolor{gray!40}$63.90$} & {\multirow{4}{*}{$5.98$M}} & {\multirow{4}{*}{$102.4$G}} \\
    {} & {\Checkmark} & {} & {$61.11$} & {$86.52$} & {$69.72$} & {} & {} \\
    {} & {} & {\Checkmark} & {$62.12$} & {$85.56$} & {$69.61$} & {} & {} \\
    {} & {\Checkmark} & {\Checkmark} & {$66.72$} & {$87.97$} & {\cellcolor{orange!60}$\mathbf{73.92}$} & {} & {} \\

    \midrule
    {\multirow{4}{*}{One2Many-STAtt}} & {} & {} & {\cellcolor{gray!40}$\mathbf{62.50}$} & {\cellcolor{gray!40}$\mathbf{85.42}$} & {\cellcolor{gray!40}$\mathbf{70.47}$} & {\multirow{4}{*}{$12.82$M}} & {\multirow{4}{*}{$106.0$G}} \\
    {} & {\Checkmark} & {} & {$65.08$} & {$87.69$} & {$71.55$} & {} & {} \\
    {} & {} & {\Checkmark} & {$65.01$} & {$86.45$} & {$70.15$} & {} & {} \\
    {} & {\Checkmark} & {\Checkmark} & {\cellcolor{orange!60}$\mathbf{68.91}$} & {\cellcolor{orange!60}$\mathbf{90.47}$} & {$73.60$} & {} & {} \\

    \bottomrule
  \end{tabular}
  \label{table:GAR}
}
\end{table*}
\section{Experiments}
\subsection{Implementation Details}
\textbf{Baselines.} Since several GAR methods are not open-sourced, we reimplement a selection of representative methods. These include skeleton-based methods COMPOSER~\cite{COMPOSER} and MPGCN~\cite{MPGCN}, as well as RGB-based methods ARG~\cite{ARG}, AT~\cite{AT}, and DIN~\cite{DIN}.
We assess their ability on the proposed SGA-INTERACT benchmark.
To fit the SGA-INTERACT benchmark, we integrate spatial-temporal modeling strategies of RGB-based methods into the One2Many framework. For the skeleton-based methods, we keep the original structure unchanged.

% The evaluation toolbox in One2Many framework will be open-sourced for the community upon publication.
For more detailed information, please refer to \cref{sec:exp-details}.

\subsection{Evaluation of GAR Task}
\label{sec:exp-GAR}
As shown in \cref{table:GAR}, we evaluate all baselines on the GAR task using several metrics: accuracy, number of model parameters, and floating point operations (FLOPs). 
The proposed One2Many-STAtt achieves the highest accuracy while maintaining a moderate number of model parameters. 
Meanwhile, several One2Many-based baselines present competitive performance, further highlighting the effectiveness of the framework.
For throughput, COMPOSER~\cite{COMPOSER} achieves extremely low computational complexity.
The One2Many framework, which processes individual sequences separately, introduces additional computational overhead.

% part discussion / conclusion
\noindent
\textbf{Discussion on Prospects}: 
To simultaneously extract scene features and leverage priors from large-scale individual action recognition data, it motivates the development of a unified backbone with compatiable training for both single and multiple person feature extraction. 
Meanwhile, despite achieving over $90\%$ $mAcc.$ on previous datasets, the baselines experience a significant performance drop.
The generally low accuracy demonstrate the challenges of SGA-INTERACT benchmark, motivating to pursue advanced spatial-temporal modeling strategies.

\subsection{Evaluation of TGAL Task}
Existing skeleton-based methods do not explicitly extract the group feature while preserving the temporal dimension.
Since the TGAL task requires temporal boundary localization in the time dimension, these skeleton-based methods are not applicable to TGAL task evaluation with the original structures unchanged.
Consequently, on SGA-INTERACT benchmark, we focus solely on evaluating methods integrated in the One2Many framework.

As shown in \cref{table:TGAL}, the proposed One2Many-STAtt demonstrates consistent superiority in the TGAL task. However, all baselines exhibit poor mAP performance. Moreover, One2Many-ARG collapses on TGAL task, obtaining with less than $1.0\%$ mAP.

\begin{table}[t]
    \centering
 {\small
\captionof{table}{Quantitative results of TGAL task on SGA-INTERACT. The best performance is highlighted in bold.}
    \begin{tabular}{l|c}
    \toprule
    {Method} & {mAP.-\%}\\
    \midrule
    {{One2Many-AT~\cite{AT}}} & {$4.01$} \\
    {{One2Many-DIN~\cite{DIN}}} & {$4.23$}\\
    {{One2Many-STAtt}} & {$\mathbf{4.52}$} \\
    \bottomrule
  \end{tabular}
  \label{table:TGAL}
}
\end{table}

\noindent
\textbf{Discussion on Prospects}: The experimental results not only reaffirm the challenges presented by SGA-INTERACT, but also illustrates evaluation limitations of GAR for group activity understanding.
It motivates further exploration of more practical group activity understanding tasks, such as TGAL for untrimmed sequences.

\subsection{Effectiveness of Augmentation}
Data augmentation is a well-established technique for enhancing model performance. However, previous studies often lack clear explanations, leading to unfair comparisons in the evaluation of group activity understanding abilities.

In \cref{table:GAR}, we first clarify and classify commonly used augmentation techniques in group activity understanding. We then separately evaluate the baselines, with and without these augmentations. Specifically, we apply flip centered on the basketball court as the spatial augmentation. For temporal augmentation, we use random frame masking of individual sequences and random frame extraction in~\cite{kwon2022context}.

All baselines benefit from the augmentation combination.
However, for MPGCN and One2Many-ARG, single type augmentation provides improvement while the performance improvement decreases after combination.
It suggests space remained to design method-specific augmentation for better group activity understanding capability.
Similar situation occurs when evaluating on TGAL task, we further discuss in ~\cref{sec:supp-TGAL-aug}.

\subsection{Effectiveness of Extra Information}
\begin{table}[t]
    \centering
 {\scriptsize
\captionof{table}{Evaluation of the impact of ball possession and team information for GAR and TGAL tasks.}
    \begin{tabular}{l|cc|c|c|c|c}
    \toprule
    {\multirow{2}{*}{Method}} & \multicolumn{2}{c}{Setting} & \multicolumn{3}{|c|}{GAR} & {TGAL}\\
    \cmidrule{2-7}
    {} & {Ball} & {Team} & {mAcc.} & {Top3-mAcc.} & {oAcc.} &  {mAP.}\\
    \midrule
    {\multirow{4}{*}{STAtt}} & {} & {} & {$62.50$} & {$85.42$} & {$70.47$} & {$4.52$} \\
    {} & {\Checkmark} & {} & {$66.67$} & {$\mathbf{90.14}$} & {$73.71$} & {$\mathbf{4.63}$}\\
    {} & {} & {\Checkmark} & {$63.14$} & {$85.14$} & {$70.26$} & {$4.36$}\\
    {} & {\Checkmark} & {\Checkmark} & {$\mathbf{68.33}$} & {$89.62$} & {$\mathbf{75.86}$} & {$4.55$} \\

    \bottomrule
  \end{tabular}
  \label{table:extra-info}
}
\end{table}

\begin{table}[t]
    \centering
 {\scriptsize
\captionof{table}{Evaluation of the impact of 3D data. Front and side view (FV, SV) indicate the projected 2D sequences as input.
$/$ denotes cross-view evaluation (i.e. FV/SV training and SV/FV evaluation).}
    \begin{tabular}{l|c|c|c}
    \toprule
    {\multirow{2}{*}{Method}} & \multirow{2}{*}{Setting} & \multicolumn{2}{c}{GAR}\\
    \cmidrule{3-4}
    {} & {} & {mAcc.} &  {oAcc.}\\
    \midrule
    {\multirow{3}{*}{COMPOSER~\cite{COMPOSER}}} & {FV 2D} & {$15.23_{/5.29}$} & {$29.96_{/21.17}$} \\
    {} & {SV 2D} & {$31.14_{/11.66}$} & {$41.81_{/6.05}$} \\
    {} & {3D} & {$47.72$} & {$58.64$} \\
    \midrule
    {\multirow{3}{*}{MPGCN~\cite{MPGCN}}} & {FV 2D} & {$50.46_{/4.77}$} & {$59.90_{/17.82}$} \\
    {} & {SV 2D} & {$19.85_{/10.70}$} & {$38.47_{/22.03}$} \\
    {} & {3D} & {$56.42$} & {$67.78$} \\
    \bottomrule
  \end{tabular}
  \label{table:3d-data}
}
\end{table}

Apart from data augmentation, the use of additional information has also led to unfair comparisons without clear explanations.
For example, in Volleyball~\cite{Volleyball}, RGB-based methods cannot directly utilize 2D ball locations, whereas skeleton-based methods leverage this information to enhance reasoning from human-object interaction.
In SGA-INTERACT, taken One2Many-STAtt as the representative model, we evaluate the impact of ball possession and team information on GAR and TGAL tasks.
As shown in \cref{table:extra-info}, both extra information positively impact the performance on GAR, highlighting further exploration on information fusion to understand challenging group activity scenarios.

\begin{table}[t]
    \centering
 {\scriptsize
\captionof{table}{Ablation study on One2Many backbone. w and w/o indicate with and without, respectively.}
    \begin{tabular}{l|c|c|c}
    \toprule
    {\multirow{2}{*}{Setting}} & \multicolumn{2}{c|}{GAR}  & {TGAL}\\
    \cmidrule{2-4}
    {} & {mAcc.} & {oAcc.} & {mAP} {}\\
    \midrule
    {One2Many-AT-w/o pretraining} & {$39.79$} & {$47.31$} & {$3.74$}\\
    {One2Many-AT-w pretraining} & {$53.50$} & {$65.63$} & {$4.01$}\\
    {\emph{Improvement}} & {\emph{+13.71}} & {\emph{+18.32}} & {\emph{+0.27}} \\
    \midrule
    {One2Many-DIN-w/o pretraining} & {$26.57$} & {$38.90$} & {$3.96$}\\
    {One2Many-DIN-w pretraining} & {$57.31$} & {$63.90$} & {$4.23$}\\
    {\emph{Improvement}} & {\emph{+28.74}} & {\emph{+25.00}} & {\emph{+0.27}} \\
    \midrule
    {One2Many-STAtt-w/o pretraining} & {$52.79$} & {$59.16$} & {$3.71$}\\
    {One2Many-STAtt-w pretraining} & {$62.50$} & {$70.47$} & {$4.52$}\\
    {\emph{Improvement}} & {\emph{+9.71}} & {\emph{+11.31}} & {\emph{+0.81}} \\
    \bottomrule
  \end{tabular}
  \label{table:ablation-one2many}
}
\end{table}

\subsection{Effectiveness of 3D Data}
\label{sec:3d-data-exp}
We project the 3D sequences of SGA-INTERACT into 2D front and side views separately to assess the impact of 3D data.
As shown in \cref{table:3d-data}, baseline methods performs better when taking 3D data as input, demonstrating that 3D information can effectively enhance group activity understanding.
Meanwhile, cross-view evaluation results reveal the fatal issue of viewpoint sensitivity in 2D training, which hinders the model's application.

\noindent
\textbf{Discussion on Prospects}: 3D/multi-view data inherently retains the advantages of single-view representations while providing additional information for research on generalization and occlusion handling.
3D datasets are widely studied for individual action understanding~\cite{shahroudy2016ntu,BABEL,sener2022assembly101,tang2023flag3d}, motivating 3D dataset construction for complex group activity.

\subsection{Effectiveness of Pretraining in One2Many}

To evaluate whether prior knowledge from individual action recognition benefits group activity understanding, we compare the performance with and without pretrained backbone weights in the One2Many framework.
As shown in \cref{table:ablation-one2many}, across AT, DIN and STAtt modeling strategies for both GAR and TGAL tasks, removing pretrained weights leads to a performance drop. This result supports the motivation behind One2Many of leveraging priors and stimulates the advancement of backbone as discussed in \cref{sec:exp-GAR}.
\section{Conclusion}
In this paper, we introduce SGA-INTERACT, the first 3D skeleton-based large-scale group activity understanding benchmark.
It establishes a challenging group activity vocabulary, incorporating rich interactions and long-term dependencies.
SGA-INTERACT supports the newly proposed TGAL task, extending group activity understanding to untrimmed sequences.
By leveraging a backbone pretrained on individual action recognition, the One2Many framework enables evaluation of existing RGB-based methods while presents competitive performance.
Extensive evaluations of baseline methods illustrate that SGA-INTERACT, along with TAGL task, serves as a challenging benchmark.
Furthermore, we outline key research directions based on SGA-INTERACT, including performance enhancement techniques, extra information fusion strategies, data formulation, and model design.

\section*{Acknowledgment}
We sincerely appreciate Zhaoyu Li, Wen Zheng, Xuedeng Zeng, Bufan Wei, and Yu Zeng for their dedicated contributions to the design and annotation of group activity in SGA-INTERACT.

{\small
\bibliographystyle{ieee_fullname}
\bibliography{main}
}

\clearpage
\setcounter{page}{1}
\maketitlesupplementary
\appendix

\section*{Contents}
\noindent
1 \hspace{0.3cm} SGA-INTERACT Details \dotfill \pageref{sec:supp1} \\
2 \hspace{0.3cm} One2Many Framework Details \dotfill \pageref{sec:supp2} \\
3 \hspace{0.3cm} Supplementary Experiments \dotfill \pageref{sec:supp3} \\
4 \hspace{0.3cm} Limitation and Other Application \dotfill \pageref{sec:supp4} \\

\begin{figure*}[t]
    \centering
    \includegraphics[width=\linewidth]{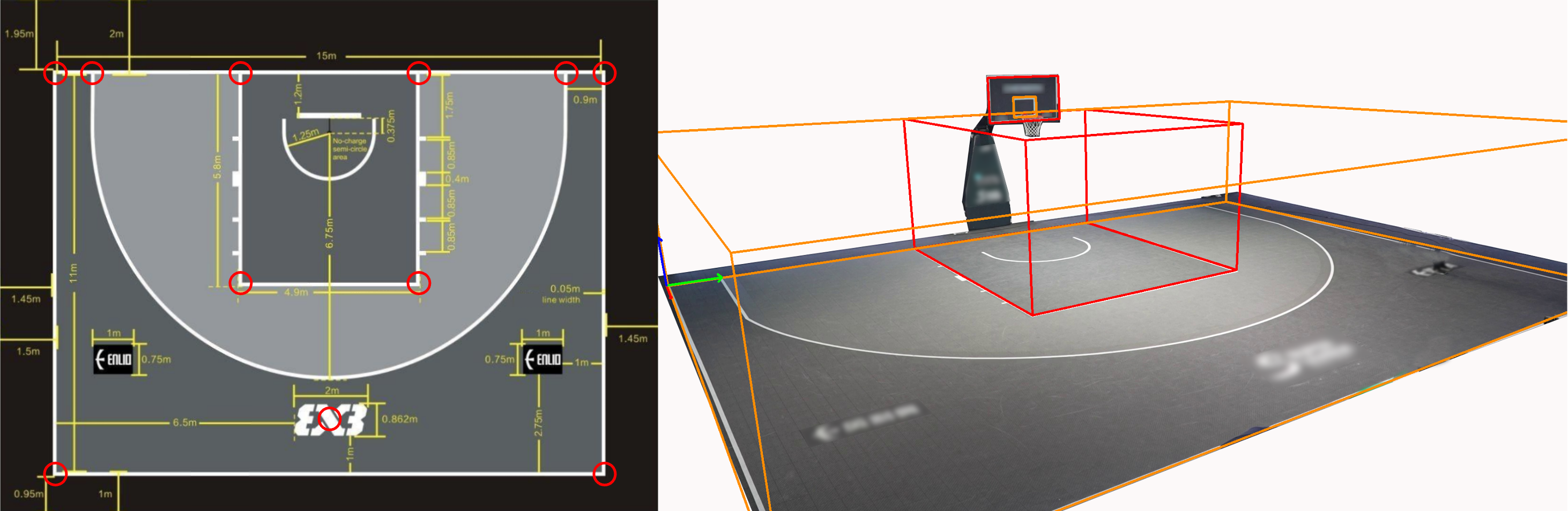}
    \caption{Demonstration of camera calibration. Left: 3$\times$3 basketball court size. Right: The 3D box is overlaid on the distorted image for visualization examination.}
    \label{fig:supp-cam-calib}
\end{figure*}

\section{SGA-INTERACT Details}
\label{sec:supp1}
\subsection{Dataset Construction Details}
\label{sec:data-cons-detail}
In this section, we focus on supplying details of the acquisition of 3D skeleton sequences. It includes raw video data collection, camera calibration, and 3D pose estimation.

\noindent
\textbf{Raw Data Collection.}
To ensure high-quality video data collection, we use four Sony ILME-FX30 cameras equipped with $18mm$ lenses for recording. It provides the ability to record for extended durations throughout the entire game day and a sufficient camera field of view to capture the game footage.
Each camera is mounted on a tripod at the corners of the basketball court, providing a static view.
To minimize occlusion-related challenges during data processing, the cameras are positioned at a height of over 3 meters from the ground.
In addition to capturing high spatial and temporal resolution at $1080p$ and $50$ FPS, we manually adjust the exposure settings to maintain consistent lighting conditions across views.

\noindent
\textbf{Camera Calibration.}
Since each camera focal length is fixed to $18mm$ for recording, we calibrate the intrinsic parameters in the laboratory before raw data collection on location.
The classical checkerboard calibration method~\cite{zhang2002flexible} is leveraged.
For the extrinsic parameters, we select $10$ keypoints on the court ground (as shown in the left of \cref{fig:supp-cam-calib}), with their 3D world coordinates defined by the International Basketball Federation (FIBA) regulations~\footnote{\scriptsize{\url{https://fiba3x3.com/docs/equipment-and-software-appendix-to-the-3x3-official-rules.pdf}}}.
For every game, we manually annotate the corresponding 2D location in one recorded image of each view. 
The extrinsic parameters are then computed based on the correspondence and examined through visualization (as demonstrated in the right of \cref{fig:supp-cam-calib}).
In addition to ground keypoints, we additionally select $9$ keypoints of the basketball for optional refinement, including the center of the basket, and corners of both the inner and outer basketball backboard boxes.
With additional height information, the intrinsic parameters and extrinsic parameters are jointly optimized.
% error
We calculate re-projection error for the extrinsic parameter calculation process. It achieves $\sim2$ pixel error in average, presenting excellent features in the application, without the need to place physical annotation tools in the real world.

\noindent
\textbf{3D Pose Estimation.}
Before applying pose estimation algorithms, we need to first synchronize multi-view videos to ensure that all views correspond to the same moment in time.
In practice, Mel-Frequency Cepstral Coefficients (MFCC)~\cite{muda2010voice} are utilized for the purpose due to their computational efficiency in audio and frame-level error minimization.
Once synchronized, the multi-view videos are segmented into game round clips based on annotations, facilitating skeleton data collection.

For multi-person multi-view 3D pose estimation, we adopt the top-down strategy.
First, we use RTMDet~\cite{lyu2022rtmdet} to detect players in each frame, followed by RTMPose~\cite{jiang2023rtmpose} for 2D keypoint estimation. Both RTMDet and RTMPose are trained on seven public datasets, and we employ pretrained weights from medium-sized models to ensure high data quality. 
The estimated keypoints are formatted according to the COCO standard~\cite{lin2014microsoft}, which includes 17 keypoints and aligns with the prevailing skeleton-based group activity recognition methods.
According to the bounding boxes provided by RTMDet, OCSORT tracker~\cite{cao2023observation} and SOLIDER Re-ID algorithm~\cite{chen2023beyond} are employed to obtain players' 2D trajectories. A database of player profile images is established to support the Re-ID process. Specifically, annotators draw $10\sim20$ tight bounding boxes for each player per game. Then player profile images are assembled and partitioned by host and guest affiliations due to different appearance of jerseys, resulting in approximately $200$ profiles images per player for Re-ID.
Since Re-ID does not guarantee correctness, multiple targets on the field may be identified as the same player. Instead of directly assigned Re-ID results, the Hungarian algorithm is used to match the target in each frame.
After obtaining player's 2D trajectories, we leverage association for multi-person triangulation to reconstruct 3D poses.
The player IDs in 3D space are determined through majority voting based on the Re-ID results across multiple views.
Additionally, a heuristic-based 3D tracking algorithms are applied, leveraging pose direction and spatial location to refine player trajectories. Player IDs are further adjusted using trajectory-based associations to enhance tracking accuracy.
Finally, based on 3D, we filter out short trajectories and out of the court trajectories.

Through this automated pipeline, approximately 0.6 million valid frames are processed efficiently. The remaining erroneous trajectories are manually corrected by two annotators within just $8$ hours, significantly reducing the time and labor required for post-processing.

\subsection{Data Distribution}
\label{sec:supp-data-dist}
In the main text, we present the distribution of group activity vocabularies in SGA-INTERACT. Here, we further analyze the spatial distribution of 3D skeleton sequences.

\begin{figure*}[t]
    \centering
    \includegraphics[width=\linewidth]{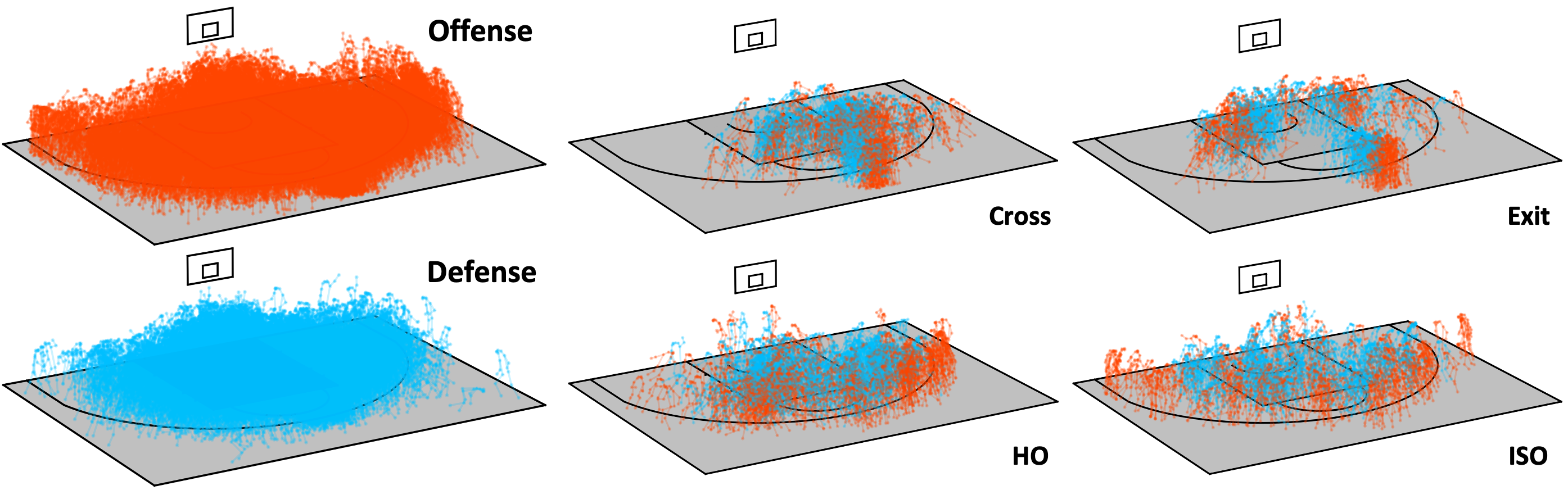}
    \caption{Visualization of 3D skeleton sequences. Host and guest team players are distinguished by orange and blue, respectively. Cross, Exit, HO and ISO indicate various group activity in SGA-INTERACT.}
    \label{fig:traj}
\end{figure*}

As shown in ~\cref{fig:traj}, the 3D trajectories of both offensive and defensive players exhibit substantial diversity, covering the entire basketball court. Moreover, for each group activity category, distinct spatial patterns emerge, ensuring the feasibility of recognition based on movement characteristics.

\subsection{Visualization}
\begin{figure*}
    \centering
    \includegraphics[width=\linewidth]{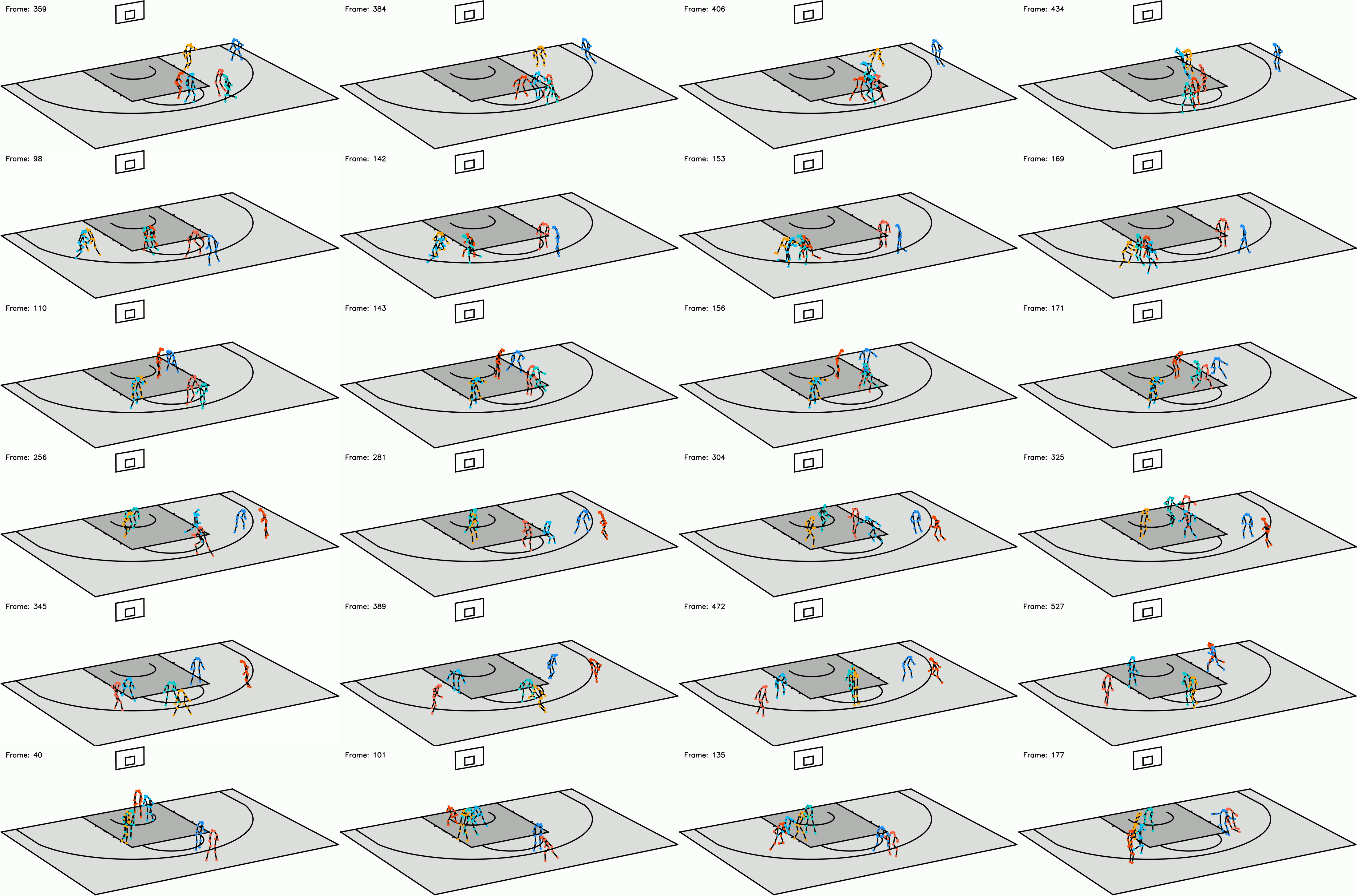}
    \caption{Group activity visualization in SGA-INTERACT. 
    Each row indicates a group activity sequence. From top to bottom, the group activities shown are PnR, DHO, Down, Basket Cut, ISO, and Exit. Best viewed when zoomed in.}
    \label{fig:supp-activity-full}
\end{figure*}

As shown in ~\cref{fig:supp-activity-full}, we select several group activity samples in SGA-INTERACT, supplying the claim of rich interaction features.

\subsection{Group Activity Vocabulary}
In \cref{tab:vocabulary}, we provide the detailed definition of each group activity category for research reference.

\begin{table*}[b]
    \centering
    \caption{Activity definition and start/end signs in SGA-INTERACT.}
    \label{tab:vocabulary}
    \scriptsize
    \begin{tabular}{c|l|l|l}
        \toprule
        {Activity} & {Definition} & {Start Sign} & {End Sign} \\
        \midrule
         {Pin} & {\makecell[l]{An off-ball player sets a positional screen for \\ another off-ball player along the sideline in the \\ lower half of the court.}} & {The player initiates movement from a stationary position.} & {\makecell[l]{Completion of ball catch or \\ Completion of screen.}} \\
         \midrule
         {Flare} & {\makecell[l]{An off-ball player uses a screen set by another \\ off-ball player to move away from the ball, \\ typically outside the three-point arc.}} & {The player initiates movement from a stationary position.} & {\makecell[l]{Completion of ball catch or \\ Completion of screen.}} \\
         \midrule
         {Reverse} & {\makecell[l]{The player being screened moves back towards the \\ ball from the far side.}} & {The player initiates movement from a stationary position.} & {\makecell[l]{Completion of ball catch or \\ Next tactical action begins}} \\
         \midrule
         {ISO} & {\makecell[l]{The ball handler completes the offense through an \\ isolation play.}} & {\makecell[l]{The player begins the offense or \\ prepares to shoot with both hands on the ball}} & {\makecell[l]{Completion of offense or \\ Making a pass.}} \\
         \midrule
         {PnR} & {\makecell[l]{An off-ball player sets a screen for the ball handler,\\ working together to complete the offense.}} & {The screener initiates movement from a stationary position.} & {\makecell[l]{Completion of offense or \\ Making a pass.}} \\
         \midrule
         {Slip} & {\makecell[l]{The screener fails to set the screen fully and \\ prematurely slips/pop out.}} & {The screener initiates movement from a stationary position.} & {\makecell[l]{Completion of ball catch or \\ Movement is terminated.}} \\
         \midrule
         {Post} & {\makecell[l]{The ball handler completes an offensive play in \\ the low post area.}} & {The low post player receives the ball.} & {Completion of offense.} \\
         \midrule
         {Pop} & {\makecell[l]{An off-ball player quickly moves from inside to \\ outside of the arc.}} & {\makecell[l]{An off-ball player inside the three-point arc \\ initiates movement from a stationary position.}} & {\makecell[l]{Completion of ball catch or \\ Movement is terminated.}} \\
         \midrule
         {Up} & {\makecell[l]{A back screen set in the upper part of the court, with \\ the player being screened moving towards the lower \\ part of the court.}} & {The screener initiates movement from a stationary position.} & {Completion of screen.} \\
         \midrule
         {Down} & {\makecell[l]{A screen set near the baseline, with the player being \\ screened moving towards the upper part of the court.}} & {The screener initiates movement from a stationary position.} & {Completion of screen.} \\
         \midrule
         {Exit} & {\makecell[l]{Two off-ball players in the paint, with one setting a \\ cross screen for the other, who uses the screen to \\ move towards the three-point arc.}} & {The player initiates movement from a stationary position.} & {\makecell[l]{Completion of ball catch or \\ Completion of screen.}} \\
         \midrule
         {HO} & {\makecell[l]{A player moves towards the ball handler and \\ completes the handoff.}} & {The player initiates movement from a stationary position.} & {Two players complete a handoff.} \\
         \midrule
         {DHO} & {\makecell[l]{A player dribbles towards another player and \\ completes the handoff.}} & {The player initiates movement from a stationary position.} & {Two players complete a handoff.} \\
         \midrule
         {FHO} & {\makecell[l]{A player moves towards the ball handler but instead \\ of receiving the ball, circling around \\ the ball handler.}} & {The player initiates movement from a stationary position.} & {Completion of movement.} \\
         \midrule
         {Keep} & {\makecell[l]{The ball handler completes a self-created offensive \\ play.}} & {The ball handler receives the ball.} & {Completion of offense.} \\
         \midrule
         {Playmaking} & {\makecell[l]{The ball handler at positions in the free-throw line, \\elbow, or low post initiates an offensive play.}} & {The ball handler receives the ball.} & {The ball handler passes the ball.} \\
         \midrule
         {Shuffle} & {\makecell[l]{An off-ball player cuts diagonally from one side \\ of the upper half of the court to the opposite corner.}} & {The player initiates movement from a stationary position.} & {\makecell[l]{Completion of ball catch or \\ Movement is terminated.}} \\
         \midrule
         {Baseline Cut} & {\makecell[l]{A off-ball player cuts along the baseline towards \\ the basket.}} & {The player initiates movement from a stationary position.} & {\makecell[l]{Completion of ball catch or \\ Movement is terminated.}} \\
         \midrule
         {Basket Cut} & {\makecell[l]{An off-ball player cuts towards the basket (not along \\ the baseline).}} & {The player initiates movement from a stationary position.} & {\makecell[l]{Completion of ball catch or \\ Movement is terminated.}} \\
         \midrule
         {Cross} & {\makecell[l]{An off-ball player uses a cross screen to move from  \\ one side of the court to the other.}} &{The player initiates movement from a stationary position.} & {\makecell[l]{Completion of ball catch or \\ Next tactical action begins.}} \\
         \midrule
         {Through} & {\makecell[l]{An off-ball player moves from one side of the court \\ to the other without using a screen.}} & {The player initiates movement from a stationary position.} & {\makecell[l]{Completion of ball catch or \\ Next tactical action begins.}} \\
         \bottomrule

    \end{tabular}
\end{table*}

\clearpage

\section{One2Many Framework Details}
\label{sec:supp2}

\subsection{STAtt Structure}
\label{sec:supp-statt-strucutre}
With the scene feature $\mathbf{\mathcal{F}}$ given, we flatten its spatial and temporal dimension into batch dimension respectively to obtain $\mathbf{\mathcal{F}}_{temp}$ and $\mathbf{\mathcal{F}}_{spatio}$. 
Defining $Atten(Q,K,V)$ as the attention mechanism~\cite{vaswani2017attention},
self-attention is first utilized to enhance representation:
\begin{equation}
    \mathbf{\mathcal{F}}_{temp}^{atten} = Atten(\mathbf{\mathcal{F}}_{temp}, \mathbf{\mathcal{F}}_{temp},\mathbf{\mathcal{F}}_{temp}),
\end{equation}
\begin{equation}
    \mathbf{\mathcal{F}}_{spatio}^{atten} = Atten(\mathbf{\mathcal{F}}_{spatio}, \mathbf{\mathcal{F}}_{spatio}, \mathbf{\mathcal{F}}_{spatio}).
\end{equation}
Then spatial-temporal and temporal-spatial correlation are established by cross-attention:
\begin{equation}
    \mathbf{\mathcal{F}}_{ST}^{atten} = Atten(Trans(\mathbf{\mathcal{F}}_{spatio}^{atten}), \mathbf{\mathcal{F}}_{temp}^{atten},
    \mathbf{\mathcal{F}}_{temp}^{atten}),
\end{equation}
\begin{equation}
    \mathbf{\mathcal{F}}_{TS}^{atten} = Atten(Trans(\mathbf{\mathcal{F}}_{temp}^{atten}), \mathbf{\mathcal{F}}_{spatio}^{atten}, , \mathbf{\mathcal{F}}_{spatio}^{atten}),
\end{equation}
where $Trans$ indicates reshape operation to align tensors for attention.
The output feature of STBlock is calculated by
\begin{equation}
    \mathbf{\mathcal{F}}^{atten} = Trans(\mathbf{\mathcal{F}}_{ST}^{atten}) + Trans( \mathbf{\mathcal{F}}_{TS}^{atten}).
\end{equation}
$\mathbf{\mathcal{F}}^{atten}$ keeps in the same shape of $\mathbf{\mathcal{F}}$. The final group feature is refined by STAtt with $N_{ST}$ stacked STBlocks.

\subsection{Task Head}
\label{sec:one2many-head}
We supply the details of the task head in One2Many framework for TGAL.
Followed by \cite{zhang2022actionformer}, the task head consists of two separate classification and regression heads, each composed of stacked 1D convolutional layers along the temporal dimension.
It generates two heatmaps: $\mathbf{H}^{cls} \in \mathbb{R}^{T_f\times N_{cls}}$ for classification and $\mathbf{H}^{reg} \in \mathbb{R}^{T_f}$ for regression.

For decoding, we adopt the strategy from CenterNet~\cite{zhou2019objects}, interpreting each heatmap entry as the probability of a classification category or a regression value.
Since the One2Many framework downsamples the time dimension from the original clip length $T$ to the group feature length $T_f$, we introduce an offset head with the same structure as the regression head to generate an offset heatmap $\mathbf{H}^{off} \in \mathbb{R}^{T_f}$.
Specifically, the decoding process first selects top-N values of $\mathbf{H}^{cls}$:
\begin{equation}
    inds = topN(\mathbf{H}^{cls}),
\end{equation}
where the category is determined by the corresponding dimension index in $N_{cls}$. 
The selected indices $inds$ serve as the temporal centers of group activity boundaries. The boundary width $time_w$ and center offset $time_o$ are then extracted as follows:
\begin{equation}
    time_w = \mathbf{H}^{reg}(inds),
\end{equation}

\begin{equation}
    time_o = \mathbf{H}^{off}(inds).
\end{equation}
Finally, the temporal boundary is represented as $[inds+time_o-time_w, inds+time_o+time_w]$.

\begin{table}[t]
    \centering
 {\small
\captionof{table}{Ablation study on One2Many-STAtt STBlocks.}
    \begin{tabular}{l|c|c|c|c}
    \toprule
    {\multirow{2}{*}{Setting}} & \multicolumn{2}{c|}{GAR}  & {TGAL} & {\multirow{2}{*}{\#Param.}}\\
    \cmidrule{2-4}
    {} & {mAcc.} & {oAcc.} & {mAP} {}\\
    \midrule
    {$N_{ST}=1$} & {$56.71$} & {$67.35$} & {$4.04$} & {$6.51$M}\\
    {$N_{ST}=2$} & {$58.59$} & {$69.29$} & {$4.03$}& {$9.67$M}\\
    {$N_{ST}=4$} & {$61.89$} & {$\mathbf{71.01}$} & {$3.89$}& {$15.98$M}\\
    {$N_{ST}=5$} & {$60.21$} & {$69.93$} & {$4.06$} & {$19.14$M}\\
    \midrule
    {STAtt ($N_{ST}=3$)} & {$\mathbf{62.50}$} & {$70.47$} & {$\mathbf{4.52}$} & {$12.82$M} \\
    \bottomrule
  \end{tabular}
  \label{table:ablation}
}
\end{table}

\section{Supplementary Experiments}
\label{sec:supp3}
\subsection{Implementation Details}
\label{sec:exp-details}
In SGA-INTERACT, each group consists of six people.
For the GAR task, we pad each sequence $T$ to a maximum length of $400$ frames, while for the TGAL task, sequences are padded to $800$ frames.
In TGAL task, the mAP is calculated across a range of $\{0.5 , 0.55, 0.6 , 0.65, 0.7 , 0.75, 0.8 , 0.85, 0.9 , 0.95\}$ tIoU thresholds.

The ST-GCN backbone is pretrained on the NTU-RGB+D dataset using the cross-subject setting. It processes input sequences of $300$ frames, which corresponds to the sliding window size, and extracts individual features with a dimension of $256$ while downsampling the temporal dimension by a factor of $4$.
For our proposed STAtt model, we use $N_{ST}=3$ STBlocks with the embedding dimension of $256$.
We optimize the model using SGD~\cite{SGD} with a momentum of $0.9$ and a weight decay of $1e^{-4}$.
For GAR and TGAL tasks, batch sizes are set to 64 and 8, respectively.
All experiments are conduction on the NVIDIA 4090 GPUs.

\begin{figure*}
    \centering
    \begin{tabular}{cc}
        % First row
        \begin{subfigure}{0.3\linewidth}
            \centering
            \includegraphics[width=\textwidth]{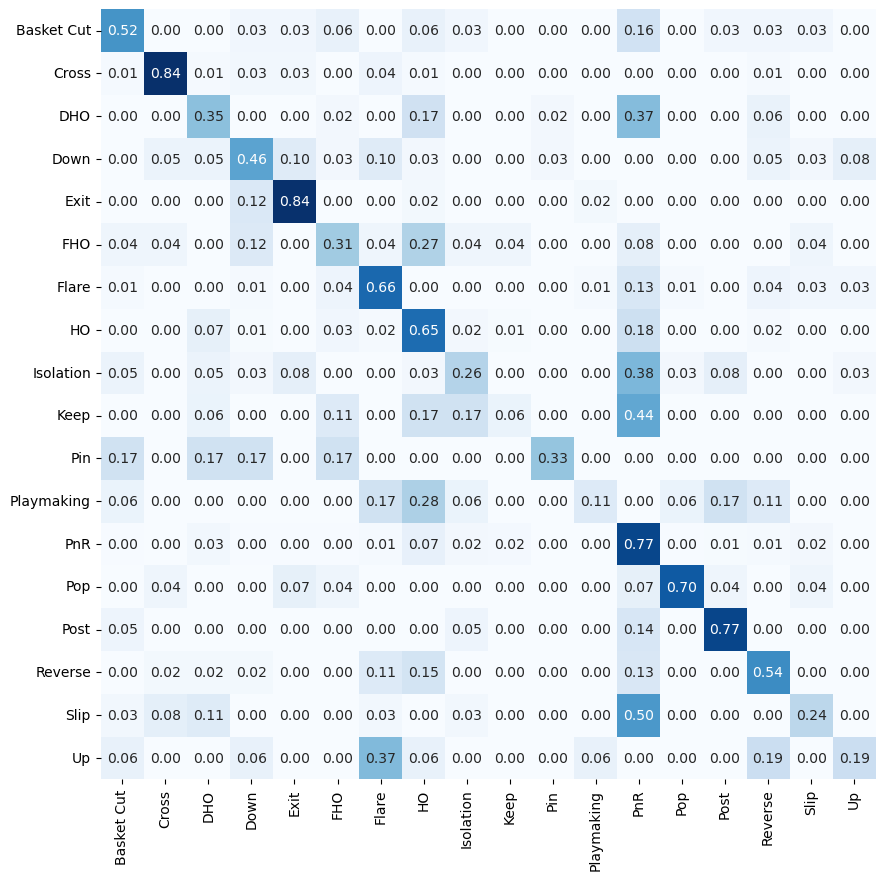}
            \caption{COMPOSER}
        \end{subfigure}
        \begin{subfigure}{0.3\linewidth}
            \centering
            \includegraphics[width=\textwidth]{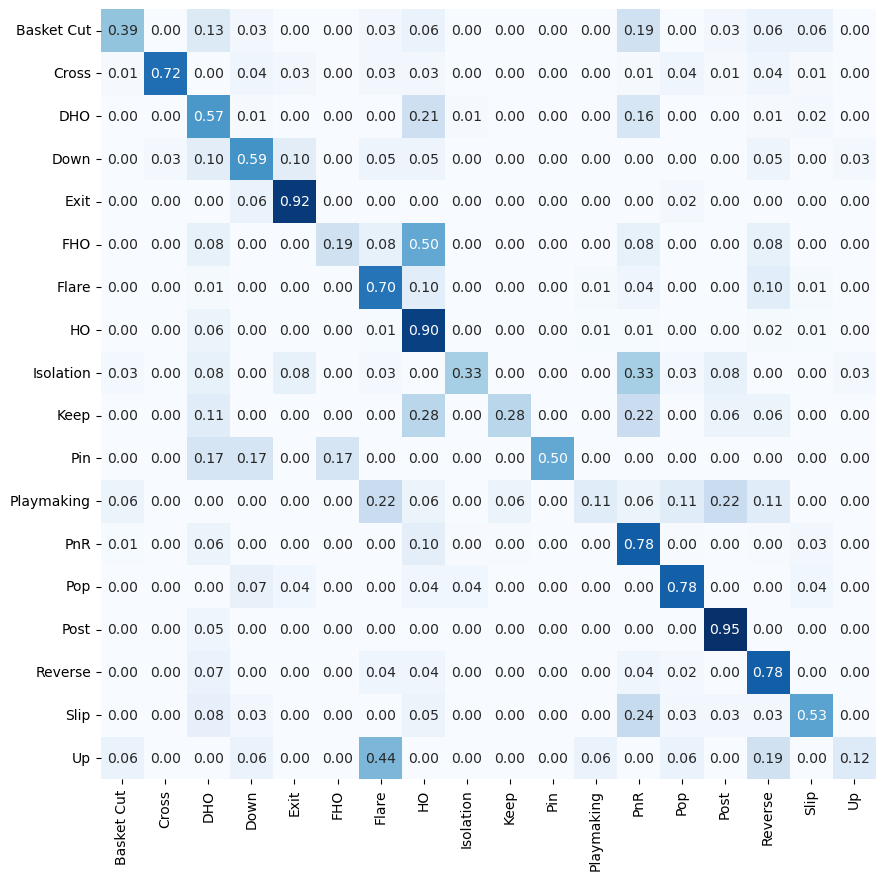}
            \caption{MPGCN}
        \end{subfigure}
        \begin{subfigure}{0.3\linewidth}
            \centering
            \includegraphics[width=\textwidth]{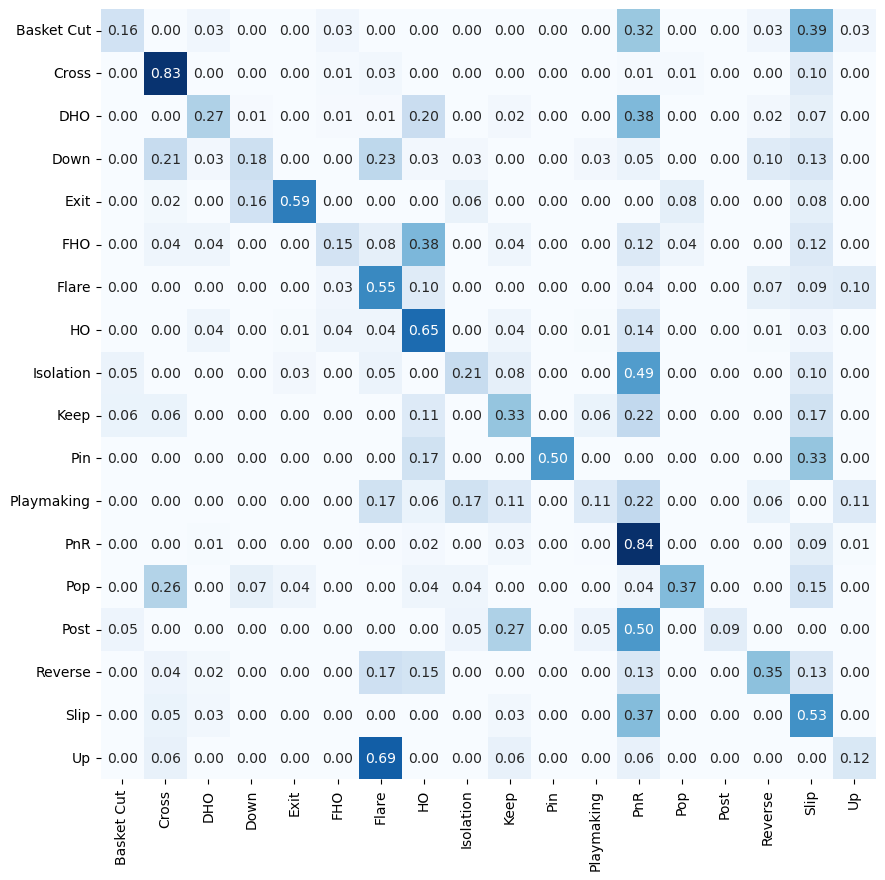}
            \caption{One2Many-ARG}
        \end{subfigure} \\
        
        % Second row
        \begin{subfigure}{0.3\linewidth}
            \centering
            \includegraphics[width=\textwidth]{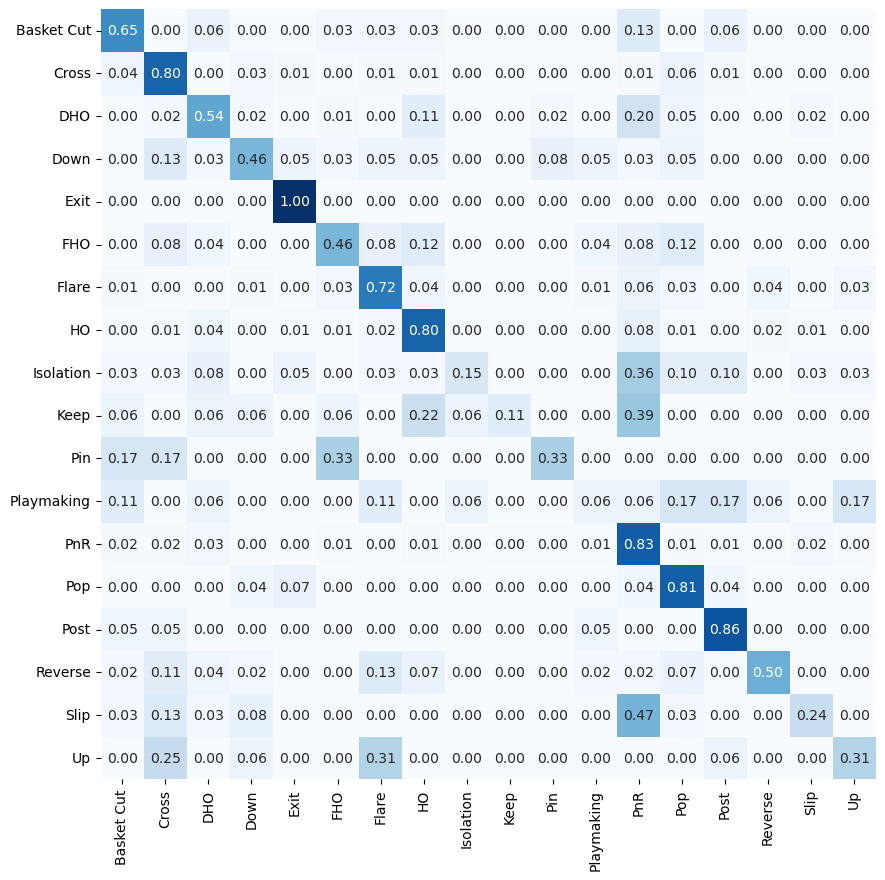}
            \caption{One2Many-AT}
        \end{subfigure}
        \begin{subfigure}{0.3\linewidth}
            \centering
            \includegraphics[width=\textwidth]{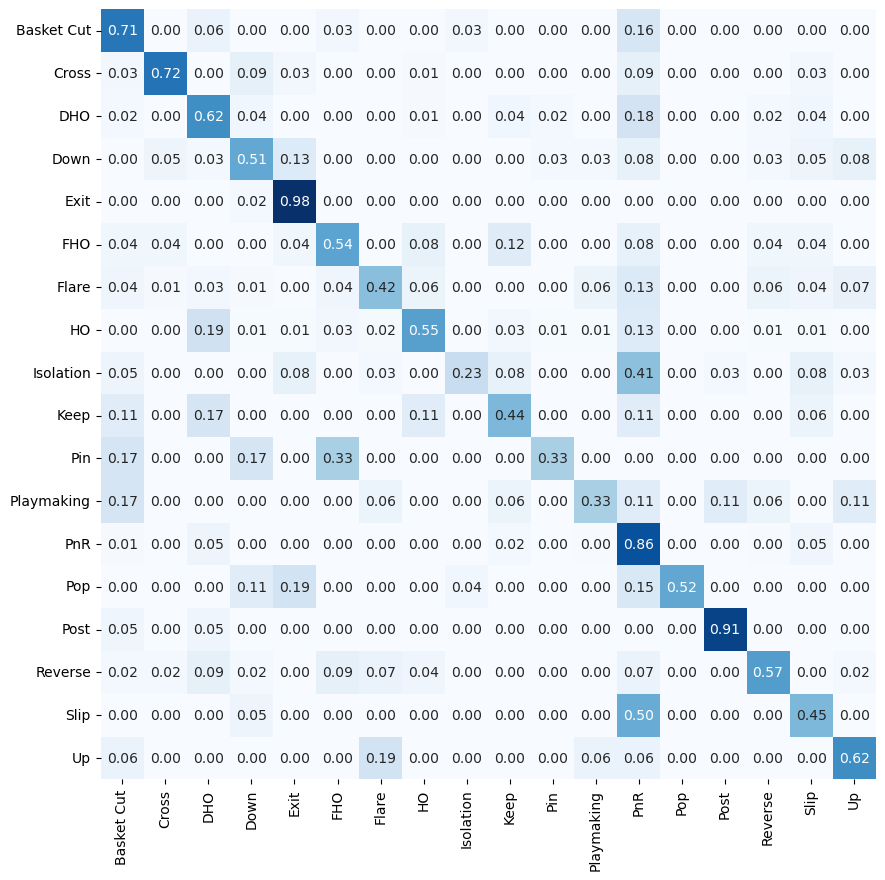}
            \caption{One2Many-DIN}
        \end{subfigure}
        \begin{subfigure}{0.3\linewidth}
            \centering
            \includegraphics[width=\textwidth]{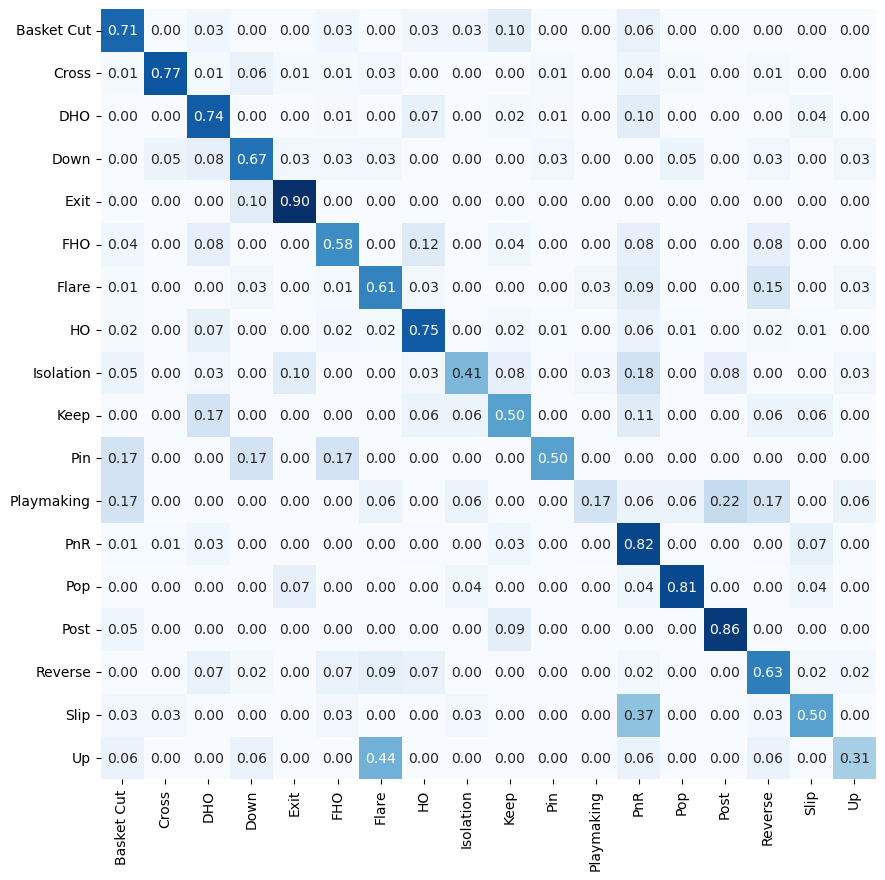}
            \caption{One2Many-STAtt}
        \end{subfigure}
    \end{tabular}

    \caption{Confusion matrices of baseline methods in GAR task. Darker color represents higher accuracy.}
    \label{fig:supp-cm}
\end{figure*}

\subsection{Ablation Study}
We conduct an ablation study on the number of STBlocks ($N_{ST}$) in the proposed STAtt model.

To evaluate its impact, we compare the performance of different $N_{ST}$ values ranging from 1 to 5 on both the GAR and TGAL tasks. As shown in \cref{table:ablation}, the model with $N_{ST} = 3$ achieves the best results across most metrics, demonstrating an optimal balance between feature extraction capacity and computational efficiency.

\subsection{In-depth Quantitative Results}
In \cref{fig:supp-cm}, we demonstrate the recognition accuracy of each category in SGA-INTERACT for baseline methods under the no-augmentation setting. The results indicate that several activities remain challenging for all baselines, highlighting the need for more advanced spatial-temporal modeling strategies.

\begin{figure}
    \centering
    \includegraphics[width=\linewidth]{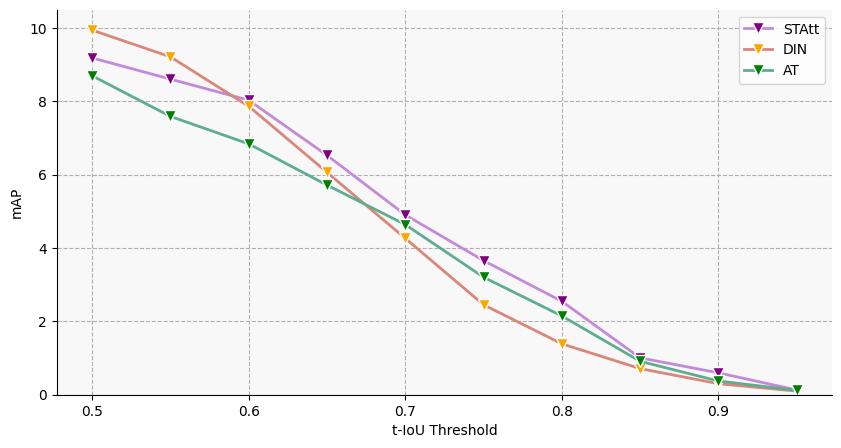}
    \caption{mAP performance of baseline methods in TGAL task.}
    \label{fig:supp-mAP}
\end{figure}

In \cref{fig:supp-mAP}, we present the mAP performance of baseline methods across various t-IoU thresholds in TGAL task. Notably, all baselines struggle to accurately localize group activities in SGA-INTERACT at high t-IoU thresholds (i.e., above $0.85$), highlighting the challenge of group activity understanding in untrimmed videos.

\begin{table}[t]
    \centering
 {\scriptsize
\captionof{table}{Quantitative results of TGAL task on SGA-INTERACT. The best performance is highlighted.}
    \begin{tabular}{l|cc|c}
    \toprule
    {Method} & {Spatial Aug.} & {Temporal Aug.} & {mAP.-\%}\\

    \midrule
    {\multirow{3}{*}{One2Many-AT~\cite{AT}}} & {} & {} & {\cellcolor{gray!40}$4.01$} \\
    {} & {\Checkmark} & {} & {$4.25$} \\
    {} & {} & {\Checkmark} & {$4.18$}\\

    \midrule
    {\multirow{3}{*}{One2Many-DIN~\cite{DIN}}} & {} & {} & {\cellcolor{gray!40}$4.23$}\\
    {} & {\Checkmark} & {} & {$3.96$} \\
    {} & {} & {\Checkmark} & {$3.84$} \\

    \midrule
    {\multirow{3}{*}{One2Many-STAtt}} & {} & {} & {\cellcolor{gray!40}$\mathbf{4.52}$} \\
    {} & {\Checkmark} & {} & {$4.25$} \\
    {} & {} & {\Checkmark} & {$3.91$} \\

    \bottomrule
  \end{tabular}
  \label{table:supp-TGAL}
}
\end{table}

\subsection{Augmentation on TGAL Task}
\label{sec:supp-TGAL-aug}
In \cref{table:supp-TGAL}, we analyze the influence of augmentation on baseline methods in TGAL task. 
Different from the results in GAR task, spatial and temporal augmentations do not introduce performance improvement in all baselines.
Instead of evading this outcome, we highlight the issue by presenting the evaluation results.
A potential explanation is that current baselines have limited group activity understanding capabilities, and augmentation introduces more complex recognition patterns, ultimately interfering with the models.
It remains a challenge in developing advanced TGAL algorithms and corresponding augmentation techniques.

\section{Limitation and Other Applications}
\label{sec:supp4}
In SGA-INTERACT benchmark, we explore the benefits of utilizing 3D information by projecting it into 2D. 
However, several potential 2D tasks derived from 3D data remain unexplored:

1) Cross view generalization. Given multiple observation viewpoints, the challenge is to train on a subset of views and evaluate the model's ability to recognize group activities in novel views.

2) Occlusion/truncation modeling.  The current projection process in \cref{sec:3d-data-exp} does not account for the effects of occlusion and truncation, as all 2D keypoints remain ``visible''. Generating occluded and truncated scenarios can enhance the model's robustness and improve its ability to handle common challenges in 2D group activity understanding.

\noindent
Based on SGA-INTERACT data, it provides scarce 3D skeleton sequences in sports along with detailed language annotations containing high-level semantics.
This dataset can facilitate research in areas such as 3D human motion synthesis and physics-based character animation.

\end{document}